\documentclass{ametsocV6.1}
\usepackage{graphicx}
\usepackage{multirow}
\usepackage{caption}
\usepackage{subcaption}
\usepackage[export]{adjustbox}
\usepackage{algorithm}
\usepackage[noend]{algpseudocode}
\algrenewcommand{\algorithmicrequire}{\textbf{Require}}
\algrenewcommand{\algorithmicensure}{\textbf{Ensure}}
\newtheorem{theorem}{Theorem}
\newtheorem{lemma}{Lemma}
\usepackage{natbib}
\newtheorem{remark}{Remark}[section]
\setcitestyle{numbers,square}
\usepackage{xcolor}

\usepackage{float}
\usepackage{booktabs}
\usepackage{hyperref}
\usepackage{cleveref}
\usepackage{bm}

\makeatletter
\let\linenumbers\relax                
\def\@LN@FinalTest{0}                 
\gdef\@LN@FinalTest{0}
\let\@LN@print\@gobble
\makeatother

\title{Hyperspectral Super-Resolution with Inter-Image Variability via Degradation-based Low-Rank and Residual Fusion Method}

\authors{Yue Wen,\aff{a} \correspondingauthor{Minru Bai, minru-bai@hnu.edu.cn} 
	Kunjing Yang,\aff{b} 
	Minru Bai,\aff{c} 
}

\affiliation{\aff{a}{School of Mathematics, Hunan University, Changsha, Hunan, China}\\
	\aff{b}{School of Mathematics, Hunan University, Changsha, Hunan, China}\\
	\aff{c}{School of Mathematics, Hunan University, Changsha, Hunan, China}\\ }

\abstract{The fusion of hyperspectral image (HSI) with multispectral image (MSI) provides an effective way to enhance the spatial resolution of HSI.
However, due to different acquisition conditions, there may exist spectral variability and spatially localized changes between HSI and MSI, referred to as inter-image variability, which can significantly affect the fusion performance.
Existing methods typically handle inter-image variability by applying direct transformations to the images themselves, which can exacerbate the ill-posedness of the fusion model.
To address this challenge, we propose a Degradation-based Low-Rank and Residual Fusion (DLRRF) model.
First, we model the spectral variability as change in the spectral degradation operator.
Second, to recover the lost spatial details caused by spatially localized changes, we decompose the target HSI into low-rank and residual components, where the latter is used to capture the lost details.
By exploiting the spectral correlation within the images, we perform dimensionality reduction on both components.
Additionally, we introduce an implicit regularizer to utilize the spatial prior information from the images.
The proposed DLRRF model is solved using the Proximal Alternating Optimization (PAO) algorithm within a Plug-and-Play (PnP) framework, where the subproblem regarding implicit regularizer is addressed by an external denoiser.
We further provide a comprehensive convergence analysis of the algorithm.
Finally, extensive numerical experiments demonstrate that DLRRF achieves superior performance in fusing HSI and MSI with inter-image variability.}

\begin{document}

\maketitle
%
%
%
%
%
%

\section{INTRODUCTION}
\label{section 1}
Hyperspectral images (HSIs) contain rich spectral information, with each pixel capturing detailed reflectance distributions across numerous spectral bands \cite{Landgrebe2002}. 
Since different materials exhibit distinct spectral signatures, the extensive spectral coverage of HSIs enables accurate material identification, which facilitates broad applications of HSIs, including classification \cite{Akhtar2018} \cite{Peng2019}, anomaly detection \cite{Kang2017}, and disease diagnosis \cite{Akbari2010}.
However, these fundamental physical constraints in imaging systems inherently impose a trade-off between spatial and spectral resolution of HSIs \cite{Shaw2003}.
As a result, HSIs typically exhibit low spatial resolution. 
In contrast, multispectral images (MSIs) provide high spatial resolution but compromise on spectral resolution. 
To overcome this limitation, a widely adopted strategy is to fuse a high spectral resolution HSI with a high spatial resolution MSI, termed HSI-MSI fusion \cite{Yokoya2017} or hyperspectral super-resolution (HSR) problem \cite{Borsoi2021}. 


Existing HSI-MSI fusion approaches can be broadly categorized into three methodologies: deep learning-based methods, matrix factorization (MF)-based methods, and tensor factorization (TF)-based methods. 
Deep learning-based methods leverage neural networks to extract intrinsic features from images and synthesize a fused image with high spatial and spectral resolution \cite{Li2022} \cite{Yao2020}. 
Palsson \textit{et al}. \cite{Palsson2017} proposed a fusion method based on a 3D convolutional neural network (CNN) combined with principal component analysis (PCA) for dimensionality reduction.
Dian \textit{et al}. \cite{Dian2018} integrated a deep learning prior into the fusion framework and achieved image reconstruction by solving a Sylvester equation.
The work in \cite{Wang2023a} developed a spatial-spectral implicit neural representation fusion network to restore the continuous spatial and spectral information of the images.
Liu \textit{et al}. \cite{Liu2024} utilized a low-rank Transformer network (LRTN) to effectively exploit both the intrinsic correlations and the global spatial correlations within images.
Nevertheless, deep learning-based approaches often exhibit poor generalization \cite{Luo2025}.
 
MF-based methods can be roughly classified into two categories: spectral unmixing-based methods and subspace representation-based methods.
Spectral unmixing-based approaches unmix the target HSI into pure endmembers matrix and abundance matrix \cite{Chen2021}. 
Yokoya \textit{et al}. \cite{Yokoya2012} introduced a coupled nonnegative matrix factorization (CNMF) method based on a linear spectral unmixing. 
Subspace representation-based methods exploit the high spectral correlation of HSIs to project the image data into low-dimensional subspace.
Simoes \textit{et al}. \cite{Simoes2014} adopted a subspace representation framework and presented how to estimate the spatial blur and spectral response in the case of the blind fusion scenario.
Dong \textit{et al}. \cite{Dong2016} proposed a joint estimation strategy for the spectral basis and sparse coefficients by leveraging the spatio-spectral sparsity inherent in the HSIs.
The work in \cite{Dian2018a} employed superpixel segmentation to oversegment MSI, thereby enforcing low-rank structure within each superpixel.
However, these MF-based methods require firstly unfolding the three-dimensional images into matrices, which will disrupt the intrinsic structure of the images.

Given that the HSIs and MSIs can be naturally represented as three-order tensors, a growing number of TF-based methods have been developed, which better leverage the multilinear features of the images \cite{Zhao2022}.
Kanatsoulis \textit{et al}. \cite{Kanatsoulis2018a} established identifiability guarantees for the target HSI by leveraging Candecomp/Parafac (CP) decomposition. 
The work in \cite{Li2018} proposed a coupled sparse tensor factorization (CSTF) method based on Tucker decomposition.
Dian \textit{et al}. \cite{Dian2019} introduced low tensor multi-rank (LTMR) regularization to simultaneously exploit high correlations among spectral bands and non-local self-similarities of target HSI.
The method in \cite{Zhang2019} established the identifiability of target HSI through block-term decomposition (BTD), meanwhile modeled the physical properties of the images.
Xu \textit{et al}. \cite{Xu2019} incorporated the nonlocal similarity, tensor dictionary learning, and tensor sparse coding together, basing on tensor product (t-product) tensor sparse representation. 
The work in \cite{Dian2019a} clustered non-local similar cubes into groups to form highly correlated four-dimensional tensors, and imposed a low tensor train rank constraint to effectively capture the correlations among spatial, spectral, and non-local dimensions.
Chen \textit{et al}. \cite{Chen2022} developed a factor-smoothed TR decomposition (FSTRD) model based on Tensor Ring (TR) decomposition, aiming to preserve the spatial–spectral continuity of the reconstructed images.
Yang \textit{et al}. \cite{Yang2025} introduced a hyperspectral blind fusion approach based on triple decomposition, designed to fully exploit the intrinsic spatial structures inherent in images.

Most existing algorithms assume that HSI and MSI data are collected under the same conditions.
In reality, however, despite the number of optical satellites orbiting the Earth (e.g. Sentinel, Orbview, Landsat and Quickbird missions) has been increasing, very few platforms are equipped with both hyperspectral and multispectral sensors \cite{Kaufmann2006} \cite{Eckardt2015}. 
Consequently, fusing HSI and MSI data acquired by separate platforms has become a mainstream method for reconstructing target HSI \cite{Hilker2009} \cite{Emelyanova2013}.  
Since these separate satellite platforms have different orbital parameters and revisit times, the HSI and MSI data are often acquired at different time instants. 
Consequently, the images can be impacted by illumination, atmospheric, seasonal changes and so on, resulting in spectral variability and spatially localized changes between the HSI and the MSI, a phenomenon referred to as inter-image variability \cite{Borsoi2021}.
Such variability adversely affects the performance of image fusion algorithms.
To address this challenge, Borsoi \textit{et al}. \cite{Borsoi2020} first introduced seasonal spectral variability, which attributed spectral variability to changes in the endmember matrix building upon the spectral unmixing of target HSI.
The work in \cite{Borsoi2021} proposed a coupled tensor approximation method that explicitly accounts for spatially and spectrally localized variations, and established theoretical recovery guarantees under mild conditions.
Prevost \textit{et al}. \cite{Prevost2022} introduced a coupled LL1 block-tensor decomposition method to handle spectral variability, and also derived exact recovery conditions for the target HSI.
The method in \cite{Fu2022} presented a group sparsity constrained fusion (GSFus) framework that incorporates a denoiser to enhance the spatial self-similarity of target HSI.
However, these methods model inter-image variability as the change of image self, which may lead to more severe model pathology.

In this paper, we propose a Degradation-based Low-Rank and Residual Fusion (DLRRF) model to address the aforementioned issue.
Specifically, since the spectral degradation operator can characterize the spectral correspondence relationship between the target HSI and the observed MSI, we model spectral variability as change in the spectral degradation operator. 
Furthermore, to recover the lost spatial details caused by spatially localized changes, we propose to partition the target HSI into a low-rank component and a residual component, where 
residual component is utilized to retrieve the lost spatial details. 
Given that the HSIs typically exhibit strong spectral correlations, subspace representation has become a common strategy for reducing dimensionality while improving estimation accuracy. 
Accordingly, we employ subspace representation method to decompose both components along the spectral dimension into their respective coefficient tensors and spectral dictionaries.
However, due to the presence of inter-image variability, some manually designed priors yield limited performance improvements in our model.
To overcome this limitation, we incorporate an implicit regularizer into the fusion framework to extract spatial prior information from the images, with the solution associated with the implicit regularizer solved by an external denoiser. 
The proposed DLRRF model is addressed via the Proximal Alternating Optimization (PAO) algorithm, with the overall optimization framework embedded within the Plug-and-Play (PnP) paradigm. 
Theoretically, we further provide a comprehensive convergence analysis of the PAO algorithm based on the Kurdyka–Łojasiewicz (KL) property, where we prove that the sequence generated by the algorithm converges to a critical point under certain conditions.

The main contributions of this paper are summarized as follows:
\begin{itemize}
	\item We propose a novel modeling strategy for spectral variability by characterizing it as change in the spectral degradation operator.
	Additionally, the target HSI is decomposed into low-rank and residual components, where the latter is utilized to restore the lost details caused by spatially localized changes. 
	To reduce computational complexity, both components are projected into low-dimensional subspaces along the spectral dimension.
	Furthermore, an implicit regularization is introduced to exploit the spatial prior information inherent in the images. 
	\item The proposed DLRRF model is solved using the PAO algorithm embedded within the PnP framework, with subproblem regarding the implicit regularization addressed by an external denoiser.
	Moreover, we provide a theoretical convergence analysis of the algorithm based on Kurdyka–Łojasiewicz (KL) property.
	\item 
	Extensive experiments on HSI-MSI fusion problem with inter-image variability demonstrate that DLRRF mothod achieves superior performance compared to some well-known methods.
	Furthermore, we apply the fused images to the image classification task, and the experimental result demonstrates that they lead to higher classification accuracy.
\end{itemize}

The remainder of this paper is organized as follows.
In Section~\ref{section 2}, we introduce the necessary notation and preliminaries, and present the fundamental formulation of the HSR problem. 
Section~\ref{section 3} proposes the DLRRF model.
The corresponding PAO algorithm is outlined in Section~\ref{section 4}, and convergence analysis of the algorithm is provided in Section~\ref{section 5}.
Section~\ref{section 6} presents extensive numerical experiments to evaluate the performance of DLRRF in the presence of inter-image variability.
Finally, we make a conclusion in Section~\ref{section 7}.

\section{PRELIMINARIES AND PROBLEM FORMULATION}
\label{section 2}
\subsection{Notation and Preliminaries}
In this paper, scalar values are denoted using lowercase letters, such as $a$; vectors are represented by boldface lowercase symbols, such as $\mathbf{a}$; matrices are expressed with uppercase bold letters, such as $\mathbf{A}$; tensors are denoted by calligraphic symbols, such as $\mathcal{A}$; and sets or number fields are shown in blackboard bold, such as $\mathbb{R}$. 
We next provide a brief introduction to tensor. 
For a more comprehensive treatment, see \cite{Kolda2009}.

A tensor of order $N$ is written as $\mathcal{A} \in \mathbb{R}^{I_1 \times I_2 \times \cdots \times I_N}$, with its $(i_1, \ldots, i_N)$-th element represented as $\mathbf{a}_{i_1, \ldots, i_N}$, where $1 \leq i_n \leq I_n$ for $n = 1, \ldots, N$.  
The Frobenius norm of $\mathcal{A}$ is defined as $\|\mathcal{A}\|_F = \sqrt{\sum \mathbf{a}_{i_1, \ldots, i_N}^2}$, and $\| \cdot \|$ also represents the Frobenius norm in this paper.
The mode-n unfolding vectors of tensor $\mathcal{A}$ are obtained from $\mathcal{A}$ by changing index $i_n$, while keeping the other indices fixed.
The mode-n unfolding matrix $\mathbf{A}_{(n)} \in \mathbb{R}^{I_n \times (I_1 I_2 \cdots I_{n-1} I_{n+1} \cdots I_N)}$ of tensor $\mathcal{A}$ is defined by arranging all the mode-n vectors as the columns of the matrix.
Specifically, for a third-order tensor $\mathcal{A} \in \mathbb{R}^{I_1 \times I_2 \times I_3}$, its mode-3 unfolding is denoted as $\mathbf{A}_{(3)} \in \mathbb{R}^{I_3 \times (I_1 I_2)}$, which we may abbreviate as $\mathbf{A} \in \mathbb{R}^{I_3 \times I_1 I_2}$.

The mode-$n$ product between a tensor $\mathcal{A} \in \mathbb{R}^{I_1 \times I_2 \times \cdots \times I_n \times \cdots \times I_N}$ and a matrix $\mathbf{B} \in \mathbb{R}^{J_n \times I_n}$ is denoted by $\mathcal{A} \times_n \mathbf{B}$, which yields an $N$-order tensor $\mathcal{M} \in \mathbb{R}^{I_1 \times I_2 \times \cdots \times J_n \times \cdots \times I_N}$. The entries of $\mathcal{M}$ are defined as:
\begin{equation}
	\mathcal{M}_{i_1, \ldots, i_{n-1}, j_n, i_{n+1}, \ldots, i_N} = \sum_{i_n} \mathbf{a}_{i_1, \ldots, i_{n-1}, i_n, i_{n+1}, \ldots, i_N} b_{j_n i_n}.
	\nonumber
\end{equation}
Equivalently, the mode-$n$ product can be expressed via matrix multiplication using the mode-$n$ unfolding $\mathbf{M}_{(n)} = \mathbf{B} \mathbf{A}_{(n)}$. Moreover, for distinct modes in a series of multiplications, the order of multiplications is irrelevant. Specifically, for $m \neq n$,
\begin{equation}
	\mathcal{A} \times_m \mathbf{B} \times_n \mathbf{C} = \mathcal{A} \times_n \mathbf{C} \times_m \mathbf{B}.
	\label{eq:mode-n}
	\nonumber
\end{equation}
When the modes of multiplications are the same, it satisfies the following associative property:
transformed into:
\begin{equation}
	\mathcal{A} \times_n \mathbf{B} \times_n \mathbf{C} = \mathcal{A} \times_n (\mathbf{C} \mathbf{B}).
	\nonumber
\end{equation}

For a given tensor $\mathcal{A} \in \mathbb{R}^{I_1 \times I_2 \times \cdots \times I_N}$ and matrices $\mathbf{D}_n \in \mathbb{R}^{J_n \times I_n}$ with $n = 1, 2, \ldots, N$, we define the tensor $\mathcal{U} \in \mathbb{R}^{J_1 \times J_2 \times \cdots \times J_N}$ as:
\begin{equation}
	\mathcal{U} = \mathcal{A} \times_1 \mathbf{D}_1 \times_2 \mathbf{D}_2 \cdots \times_N \mathbf{D}_N,
	\nonumber
\end{equation}
then its vectorized form satisfies the relation:
\begin{equation}
	\mathbf{u} = (\mathbf{D}_N \otimes \mathbf{D}_{N-1} \otimes \cdots \otimes \mathbf{D}_1) \mathbf{a},
	\nonumber
\end{equation}
where `$\otimes$' denotes the Kronecker product, and $\mathbf{u} := \text{vec}(\mathcal{U}) \in \mathbb{R}^{J}$ with $J = \prod_{n=1}^N J_n$ and $\mathbf{a} := \text{vec}(\mathcal{A}) \in \mathbb{R}^{I}$ with $I = \prod_{n=1}^N I_n$ are vectors obtained by stacking all the mode-1 vectors of the tensors $\mathcal{U}$ and $\mathcal{A}$, respectively. 

\subsection{Problem Formulation}
Let $\mathcal{Y} \in \mathbb{R}^{w \times h \times S}$ represent a low spatial resolution HSI, where $w$ and $h$ are the spatial dimensions, and $S$ is the number of spectral bands.
Let $\mathcal{Z} \in \mathbb{R}^{W \times H \times s}$ represent a high spatial resolution MSI, with $W > w$, $H > h$ as the spatial dimensions and $s < S$ as the number of spectral bands.

The observed low spatial resolution HSI $\mathcal{Y}$ can be regarded as the spatially degraded version of the target high spatial resolution HSI (HR-HSI) $\mathcal{X} \in \mathbb{R}^{W \times H \times S}$, which can be formulated as:
\begin{equation}
	\mathcal{Y} = \mathcal{X} \times_1 \mathbf{P}_1 \times_2 \mathbf{P}_2 + \mathcal{N}_h,
	\label{HSI}
\end{equation}
where $\mathcal{N}_h \in \mathbb{R}^{w \times h \times S}$ denotes the additive Gaussian noise, and $\mathbf{P}_1 \in \mathbb{R}^{w \times W}$ and $\mathbf{P}_2 \in \mathbb{R}^{h \times H}$ are the downsampling matrices along the width and height dimensions, respectively. 
Specifically, the model in equation \eqref{HSI} is based on the separability assumption \cite{Rivenson2009}.

The observed high spatial resolution MSI $\mathcal{Z}$ can be regarded as the spectrally degraded version of $\mathcal{X}$, which can be written as:
\begin{equation}
	\mathcal{Z} = \mathcal{X} \times_3 \mathbf{P}_3 + \mathcal{N}_m,
	\label{MSI}
\end{equation}
where $\mathbf{P}_3 \in \mathbb{R}^{s \times S}$ is the spectral response function (SRF) matrix, and $\mathcal{N}_m \in \mathbb{R}^{W \times H \times s}$ represents the additive Gaussian noise.

If the HSIs and MSIs are acquired under the identical conditions, both equations \eqref{HSI} and \eqref{MSI} can hold simultaneously.
However, in practical applications, HSI and MSI data are often captured by different times, leading to inter-image variability \cite{Borsoi2021}.
This variability manifests in two primary forms: spectral variability and spatially localized changes.
Spectral variability arises from illumination, atmospheric or seasonal changes, and can appear even when the time interval between acquisitions is relatively short \cite{Borsoi2020}. 
Spatially localized changes occur when certain regions of the scene undergo differential seasonal effects or abrupt alterations, such as the appearance or disappearance of objects \cite{Liu2019}.
Consequently, owing to such inter-image variability, the equations \eqref{HSI} and \eqref{MSI} may no longer accurately describe the observation process.

\section{THE PROPOSED DLRRF MODEL}
\label{section 3}
In this section, we propose a Degradation-based Low-Rank and Residual Fusion (DLRRF) model to address the issues mentioned above.
Specifically, since the spectral degradation operator $\mathbf{P}_3$ can be  employed to model the spectral correspondence relationship between the target HSI and the observed MSI. 
Consequently, spectral variability can be effectively characterized by variations in this operator, reflecting changes in the spectral response characteristics across different acquisition conditions, which can be written as:
\begin{equation}
	\mathbf{P}_3 = \mathbf{R} + \Delta \mathbf{R},
	\label{3.a.delta_R}
\end{equation}
where $\mathbf{R} \in \mathbb{R}^{s \times S}$ denotes the spectral response matrix and $\Delta \mathbf{R} \in \mathbb{R}^{s \times S}$ represents the deviation accounting for spectral variability.
Moreover, spatially localized changes can introduce discrepancies in spatial content between observed HSI and MSI, potentially resulting in the loss of critical spatial detail of target HSI.
Therefore, we partition the target HSI into a low-rank component and a residual component, where the residual component is used to retrieve the lost spatial details. 
This decomposition can be formulated as:
\begin{equation}
	\mathcal{X} = \widetilde{\mathcal{L}} + \widetilde{\mathcal{E}},
    \label{XLE}
\end{equation}
where $\widetilde{\mathcal{L}}$ and $\widetilde{\mathcal{E}}$ represent the low-rank component and residual component, respectively.

To reduce computational complexity, we adopt subspace representation approach to model the low-rank component and residual component.
Specifically, each component is decomposed into the corresponding coefficient tensors and dictionaries along the spectral dimension, that is:
\begin{equation}
	\widetilde{\mathcal{L}} = \mathcal{L} \times_3 \mathbf{D}_{\mathcal{L}}, ~
	\widetilde{\mathcal{E}} = \mathcal{E} \times_3 \mathbf{D}_{\mathcal{E}},
	\label{LE}
\end{equation}
where $\mathcal{L} \in \mathbb{R}^{W \times H \times S_1}$ and $\mathcal{E} \in \mathbb{R}^{W \times H \times S_2}$ are the respective coefficient tensors, $\mathbf{D}_{\mathcal{L}} \in \mathbb{R}^{S \times S_1}$ and $\mathbf{D}_{\mathcal{E}} \in \mathbb{R}^{S \times S_2}$ represent the spectral dictionaries associated with $\widetilde{\mathcal{L}}$ and $\widetilde{\mathcal{E}}$, respectively. 
The parameters $S_1$ and $S_2$ denote the subspace dimensions spanned by the columns of $\mathbf{D}_{\mathcal{L}}$ and $\mathbf{D}_{\mathcal{E}}$, respectively.

Given that both HSI and HR-HSI possess high spectral resolution, we assume that they share common spectral subspaces. 
Furthermore, we assume that the subspaces of $\widetilde{\mathcal{L}}$ and $\widetilde{\mathcal{E}}$ are orthogonal. 
Under these assumptions, the spectral dictionaries $\mathbf{D}_{\mathcal{L}}$ and $\mathbf{D}_{\mathcal{E}}$ can be estimated from the HSI via truncated singular value decomposition (SVD), which is given by:
\begin{equation}
	\mathbf{Y} = \mathbf{U}  \mathbf{\Sigma} \mathbf{V}^T , ~
	\mathbf{D}_{\mathcal{L}} = \mathbf{U}(: , 1:S_1) , ~
	\mathbf{D}_{\mathcal{E}} = \mathbf{U}(: , S_1+1 : S_1+S_2)~(S_1 + S_2 \leq S),
	\nonumber
\end{equation}
where $\mathbf{Y}$ represents the mode-3 unfolding matrix of the tensor $\mathcal{Y}$, $\mathbf{\Sigma} \in \mathbb{R}^{S \times wh}$ contains the ordered singular values, and $\mathbf{U} \in \mathbb{R}^{S \times S}$ and $\mathbf{V} \in \mathbb{R}^{wh \times wh}$ are the left and right singular matrices, respectively.

Combining the equations \eqref{HSI}, \eqref{MSI}, \eqref{3.a.delta_R}, \eqref{XLE}, and \eqref{LE}, we formulate the proposed DLRRF model as the following optimization problem:
\begin{equation}
	\begin{aligned}
		\min_{\mathcal{L}, \mathcal{E}, \Delta \mathbf{R}} ~
		& f(\mathcal{L}, \mathcal{E}, \Delta \mathbf{R}) =
		\left\| \mathcal{L} \times_1 \mathbf{P}_1 \times_2 \mathbf{P}_2 \times_3 \mathbf{D}_{\mathcal{L}} + \mathcal{E} \times_1 \mathbf{P}_1 \times_2 \mathbf{P}_2 \times_3 \mathbf{D}_{\mathcal{E}} - \mathcal{Y} \right\|_F^2 \\
		& + \tau \left\| \mathcal{L} \times_3 (\mathbf{R}+\Delta \mathbf{R}) \mathbf{D}_{\mathcal{L}} + \mathcal{E} \times_3 (\mathbf{R}+\Delta \mathbf{R}) \mathbf{D}_{\mathcal{E}} - \mathcal{Z} \right\|_F^2
		+ \lambda \phi(\mathcal{L}),
		\label{optimization formulation}
	\end{aligned}
\end{equation}
where $\phi(\cdot)$ is a continuous and semi-algebraic function, which implicitly utilizes the spatial correlation of coefficient tensor $\mathcal{L}$. 
The parameters  $\tau > 0$, $\lambda > 0$ are regularization parameters that balance the data fidelity term and the prior term on $\mathcal{L}$, respectively.

\begin{remark} 
	Although various handcrafted priors have been proposed for coefficient tensors estimation in hyperspectral image processing, such as sparse representation \cite{Kawakami2011} \cite{Wei2015}, nonlocal spatial similarity \cite{Dong2016} \cite{Dian2019}, and spatial smoothness constraints \cite{Simoes2014} \cite{Yuan2020}, these priors do not yield significant performance improvements in our model, which may result from the presence of inter-image variability.
    To address this limitation, we employ a plugged prior as the implicit regularizer on $\mathcal{L}$, which circumvents the need for explicitly defining the regularization function $\phi(\cdot)$. 
    In this paper, we leverages an external denoiser to exploit the spatial information present within the image, without requiring a closed-form expression of the regularizer.
\end{remark}

\section{THE PAO ALGORITHM FOR SOLVING DLRRF MODEL}
\label{section 4}
In this section, we employ the Proximal Alternating Optimization (PAO) algorithm to solve the optimization model \eqref{optimization formulation}. 
The iterative updates for variables $\mathcal{L}$, $\mathcal{E}$, and $\Delta \mathbf{R}$ are explicitly given by:
\begin{equation}
	\mathcal{L}^{k+1} = \arg \min_{\mathcal{L}} 
	f(\mathcal{L}, \mathcal{E}^k, \Delta \mathbf{R}^k) 
	+ \frac{\eta}{2} \| \mathcal{L} - \mathcal{L}^k \|_F^2,
	\label{PAO L}
\end{equation}

\begin{equation}
	\mathcal{E}^{k+1} = \arg \min_{\mathcal{E}} 
	f(\mathcal{L}^{k+1}, \mathcal{E}, \Delta \mathbf{R}^k) 
	+ \frac{\eta}{2} \| \mathcal{E} - \mathcal{E}^k \|_F^2,
	\label{PAO E}
\end{equation}

\begin{equation}
	\Delta \mathbf{R}^{k+1} = \arg \min_{\Delta \mathbf{R}} 
	f(\mathcal{L}^{k+1}, \mathcal{E}^{k+1}, \Delta \mathbf{R}) + \frac{\eta}{2} \| \Delta \mathbf{R} - \Delta \mathbf{R}^k \|_F^2,
	\label{PAO DR}
\end{equation}
where $\eta > 0$ is the proximal parameter. 
The steps of PAO algorithm are summarized in Algorithm~\ref{alg:PAO}. 
In the following, we present the optimizations for each variable in detail.

\begin{algorithm}
	\caption{PAO algorithm for solving DLRRF model \eqref{optimization formulation}}
	\label{alg:PAO}
	{\bf Input:} $\mathcal{Y}, \mathcal{Z}, \mathbf{P}_1, \mathbf{P}_2, \mathbf{R}, \mathbf{D}_{\mathcal{L}}, \mathbf{D}_{\mathcal{E}}, \tau, \lambda, \eta, \mu$. 
	
	\textbf{Initialize:} $\mathcal{L}^0$, $\mathcal{E}^0$, $\Delta \mathbf{R}^0$. Set $k = 0$. 
	
	\textbf{Step1:} Compute $\mathcal{L}^{k+1}$ by solving the subproblem \eqref{PAO L};
	
	\textbf{Step2:} Compute $\mathcal{E}^{k+1}$ by solving the subproblem \eqref{PAO E};
	
	\textbf{Step3:} Compute $\Delta \mathbf{R}^{k+1}$ by solving the subproblem \eqref{PAO DR};
	
	\textbf{Step4:} If converge, the algorithm terminates; 
	
	\hspace{34pt} otherwise, set $k = k + 1$ and go to Step 1;
	
	{\bf Output:} $\mathcal{X} = \mathcal{L}^{k+1} \times_3 \mathbf{D}_{\mathcal{L}} + \mathcal{E}^{k+1} \times_3 \mathbf{D}_{\mathcal{E}}$.
\end{algorithm}

\subsubsection{Optimization with respect to $\mathcal{L}$}
With $\mathcal{E}$ and $\Delta \mathbf{R}$ fixed, the optimization problem \eqref{PAO L} becomes:
\begin{equation}
	\begin{aligned}
		\min_{\mathcal{L}} ~
		& \| \mathcal{L}  \times_1 \mathbf{P}_1 \times_2 \mathbf{P}_2 \times_3 \mathbf{D}_{\mathcal{L}} + \mathcal{E}  \times_1 \mathbf{P}_1 \times_2 \mathbf{P}_2 \times_3 \mathbf{D}_{\mathcal{E}} - \mathcal{Y} \|_F^2  \\
		&+ \tau \| \mathcal{L} \times_3 (\mathbf{R}+\bigtriangleup \mathbf{R}) \mathbf{D}_{\mathcal{L}} + \mathcal{E} \times_3 (\mathbf{R}+\bigtriangleup \mathbf{R}) \mathbf{D}_{\mathcal{E}} - \mathcal{Z} \|_F^2 \\
		&+ \lambda \phi(\mathcal{L})+\frac{\eta}{2} \| \mathcal{L} - \mathcal{L}^k \|_F^2. 
		\label{L 1}
	\end{aligned}
\end{equation}
Since directly solving \eqref{L 1} is challenging, we introduce two auxiliary variables $\mathcal{A} = \mathcal{L}$ and $\mathcal{B} = \mathcal{L}$, and then penalize these constraints in the objective function, which reformulates the original problem into the following form:
\begin{equation}
	\begin{aligned}
		\min_{\mathcal{L},\mathcal{A},\mathcal{B}} ~
		& \| \mathcal{L}  \times_1 \mathbf{P}_1 \times_2 \mathbf{P}_2 \times_3 \mathbf{D}_{\mathcal{L}} + \mathcal{E}  \times_1 \mathbf{P}_1 \times_2 \mathbf{P}_2 \times_3 \mathbf{D}_{\mathcal{E}} - \mathcal{Y} \|_F^2  \\
		&+ \tau \| \mathcal{A} \times_3 (\mathbf{R}+\bigtriangleup \mathbf{R}) \mathbf{D}_{\mathcal{L}} + \mathcal{E} \times_3 (\mathbf{R}+\bigtriangleup \mathbf{R}) \mathbf{D}_{\mathcal{E}} - \mathcal{Z} \|_F^2 \\
		&+ \lambda \phi(\mathcal{B})+\frac{\eta}{2} \| \mathcal{L} - \mathcal{L}^k \|_F^2+\frac{\mu}{2} \| \mathcal{A}-\mathcal{L} \|_F^2 +\frac{\mu}{2} \| \mathcal{B}-\mathcal{L} \|_F^2 ,
		\label{L 2}
	\end{aligned}
\end{equation}
where $\mu$ is a positive penalty parameter.
Subsequently, we solve this subproblem by alternating minimization over $\mathcal{L}$, $\mathcal{A}$, and $\mathcal{B}$.

%
%
%
%
%
%
%

\textbf{a.} The update for $\mathcal{L}$ involves minimizing~\eqref{L 2} with respect to $\mathcal{L}$, i.e.,
\begin{equation}
	\begin{aligned}
		\min_{\mathcal{L}}  ~ 
		& \| \mathcal{L}  \times_1 \mathbf{P}_1 \times_2 \mathbf{P}_2 \times_3 \mathbf{D}_{\mathcal{L}} + \mathcal{E}  \times_1 \mathbf{P}_1 \times_2 \mathbf{P}_2 \times_3 \mathbf{D}_{\mathcal{E}} - \mathcal{Y} \|_F^2  \\
		& +\frac{\eta}{2} \| \mathcal{L} - \mathcal{L}^k \|_F^2+\frac{\mu}{2} \| \mathcal{A}-\mathcal{L} \|_F^2 +\frac{\mu}{2} \| \mathcal{B}-\mathcal{L} \|_F^2.
		\label{L_T}
	\end{aligned}
\end{equation}
This problem can be equivalently rewritten as:
\begin{equation}
	\min_{\mathcal{L}} ~  \| \mathcal{L}  \times_1 \mathbf{P}_1 \times_2 \mathbf{P}_2 \times_3 \mathbf{D}_{\mathcal{L}} - \mathcal{Y}_1  \|_F^2  
	+ \xi_1 \| \mathcal{L} - \mathcal{M}_1 \|_F^2,
	\label{L_L}
\end{equation}
where $\mathcal{Y}_1=\mathcal{Y}-\mathcal{E}  \times_1 \mathbf{P}_1 \times_2 \mathbf{P}_2 \times_3 \mathbf{D}_{\mathcal{E}}$, $\displaystyle \xi_1=\frac{\eta+2 \mu}{2}$, and $\displaystyle \mathcal{M}_1=\frac{\eta \mathcal{L}^k+\mu \mathcal{A}+\mu \mathcal{B}}{\eta+2 \mu}$. Problem~\eqref{L_L} admits a closed-form solution via the method proposed in~\cite{Li2018}.

\textbf{b.} The update for $\mathcal{A}$ involves minimizing~\eqref{L 2} with respect to $\mathcal{A}$, i.e.,
\begin{equation}
	\min_{\mathcal{A}}  ~ \tau \| \mathcal{A} \times_3 (\mathbf{R}+\bigtriangleup \mathbf{R}) \mathbf{D}_{\mathcal{L}} + \mathcal{E} \times_3 (\mathbf{R}+\bigtriangleup \mathbf{R}) \mathbf{D}_{\mathcal{E}} - \mathcal{Z} \|_F^2
	+ \frac{\mu}{2} \| \mathcal{A} - \mathcal{L} \|_F^2 . 
	\label{L_A}
\end{equation}
By applying mode-3 matrix unfolding, this problem can be transformed into the following equivalent matrix form:
\begin{equation}
	\min_{\mathbf{A}} ~  \tau \|  (\mathbf{R}+\bigtriangleup \mathbf{R}) \mathbf{D}_{\mathcal{L}} \mathbf{A} +  (\mathbf{R}+\bigtriangleup \mathbf{R}) \mathbf{D}_{\mathcal{E}} \mathbf{E} - \mathbf{Z} \|_F^2 + \frac{\mu}{2} \| \mathbf{A} - \mathbf{L} \|_F^2 ,
	\label{L_AM}
\end{equation}
where $\mathbf{A}$, $\mathbf{L}$, $\mathbf{E}$ and $\mathbf{Z}$ denote the mode-3 unfolding matrices of tensors $\mathcal{A}$, $\mathcal{L}$, $\mathcal{E}$, and $\mathcal{Z}$, respectively.
According to optimality condition of \eqref{L_AM}, we get:
\begin{equation}
		[2 \tau \left((\mathbf{R} + \bigtriangleup \mathbf{R}) \mathbf{D}_{\mathcal{L}}\right)^T (\mathbf{R} + \bigtriangleup \mathbf{R}) \mathbf{D}_{\mathcal{L}} + \mu \mathbf{I} ] \mathbf{A} = \\
		2 \tau \left((\mathbf{R} + \bigtriangleup \mathbf{R}) \mathbf{D}_{\mathcal{L}}\right)^T \left(\mathbf{Z} - (\mathbf{R} + \bigtriangleup \mathbf{R}) \mathbf{D}_{\mathcal{E}} \mathbf{E} \right) + \mu \mathbf{L}.
		\label{L_A_H}
		\nonumber
\end{equation}
Define
\begin{equation}
	\begin{aligned}
		\mathbf{\Omega}_1 &= 2 \tau \left((\mathbf{R} + \bigtriangleup \mathbf{R}) \mathbf{D}_{\mathcal{L}}\right)^T (\mathbf{R} + \bigtriangleup \mathbf{R}) \mathbf{D}_{\mathcal{L}} + \mu \mathbf{I}, \\
		\mathbf{\Omega}_2 &= 2 \tau \left((\mathbf{R} + \bigtriangleup \mathbf{R}) \mathbf{D}_{\mathcal{L}}\right)^T \left(\mathbf{Z} - (\mathbf{R} + \bigtriangleup \mathbf{R}) \mathbf{D}_{\mathcal{E}} \mathbf{E} \right) + \mu \mathbf{L},
		\nonumber
	\end{aligned}
\end{equation}
then the optimal solution is given by:
\begin{equation}
	\mathbf{A}^* = {\mathbf{\Omega}_1}^{-1} ~ \mathbf{\Omega}_2.
	\label{L_A_S}
	\nonumber
\end{equation}

\textbf{c.} The update for $\mathcal{B}$ involves minimizing~\eqref{L 2} with respect to $\mathcal{B}$, i.e.,
\begin{equation}
	\min_{\mathcal{B}} ~ \lambda \phi(\mathcal{B}) + \frac{\mu}{2} \| \mathcal{B} - \mathcal{L} \|_F^2.
	\label{L_B1}
\end{equation}
Within the PnP framework, the minimizer of \eqref{L_B1} is given by a denoising operator:
\begin{equation}
	\mathcal{B} \leftarrow \text{Denoiser}\left( \mathcal{L}; \frac{\lambda}{\mu} \right),
	\label{L_B2}
	\nonumber
\end{equation}
where $\text{Denoiser}(\cdot)$ is an image denoising operator, which can be weighted nuclear norm minimization~\cite{Gu2014}, convolutional neural network-based approaches~\cite{Zhang2017} \cite{Zhang2017a}, and so on. 
In this work, we employ BM4D~\cite{Dabov2007} as the denoiser, owing to its demonstrated effectiveness in image restoration.

\subsubsection{Optimization with respect to $\mathcal{E}$}
With $\mathcal{L}$ and $\Delta \mathbf{R}$ fixed, the optimization problem \eqref{PAO E} becomes:
\begin{equation}
	\begin{aligned}
		\min_{\mathcal{E}} ~
		& \| \mathcal{L}  \times_1 \mathbf{P}_1 \times_2 \mathbf{P}_2 \times_3 \mathbf{D}_{\mathcal{L}} + \mathcal{E}  \times_1 \mathbf{P}_1 \times_2 \mathbf{P}_2 \times_3 \mathbf{D}_{\mathcal{E}} - \mathcal{Y} \|_F^2  \\
		&+ \tau \| \mathcal{L} \times_3 (\mathbf{R}+\bigtriangleup \mathbf{R}) \mathbf{D}_{\mathcal{L}} + \mathcal{E} \times_3 (\mathbf{R}+\bigtriangleup \mathbf{R}) \mathbf{D}_{\mathcal{E}} - \mathcal{Z} \|_F^2 +\frac{\eta}{2} \| \mathcal{E} - \mathcal{E}^k \|_F^2. 
		\label{E 1}
	\end{aligned}
\end{equation}
In the same way, we introduce an auxiliary variable $\mathcal{C} = \mathcal{E}$ and penalize it in the objective function, which reformulates \eqref{E 1} into:
\begin{equation}
	\begin{aligned}
		\min_{\mathcal{E},\mathcal{C}} ~
		& \| \mathcal{L}  \times_1 \mathbf{P}_1 \times_2 \mathbf{P}_2 \times_3 \mathbf{D}_{\mathcal{L}} + \mathcal{E}  \times_1 \mathbf{P}_1 \times_2 \mathbf{P}_2 \times_3 \mathbf{D}_{\mathcal{E}} - \mathcal{Y} \|_F^2  \\
		&+ \tau \| \mathcal{L} \times_3 (\mathbf{R}+\bigtriangleup \mathbf{R}) \mathbf{D}_{\mathcal{L}} + \mathcal{C} \times_3 (\mathbf{R}+\bigtriangleup \mathbf{R}) \mathbf{D}_{\mathcal{E}} - \mathcal{Z} \|_F^2 \\
		&+ \frac{\eta}{2} \| \mathcal{E} - \mathcal{E}^k \|_F^2+\frac{\mu}{2} \| \mathcal{C}-\mathcal{E} \|_F^2 ,
		\label{E 2}
	\end{aligned}
\end{equation}
where $\mu$ is a positive penalty parameter. The detailed steps for solving $\mathcal{E}$ and $\mathcal{C}$ are presented below.

%
%
%
%
%
%

\textbf{a.}  The update for $\mathcal{E}$ involves minimizing~\eqref{E 2} with respect to $\mathcal{E}$, i.e.,
\begin{equation}
	\begin{aligned}
		\min_{\mathcal{E}}  ~
		& \| \mathcal{L}  \times_1 \mathbf{P}_1 \times_2 \mathbf{P}_2 \times_3 \mathbf{D}_{\mathcal{L}} + \mathcal{E}  \times_1 \mathbf{P}_1 \times_2 \mathbf{P}_2 \times_3 \mathbf{D}_{\mathcal{E}} - \mathcal{Y} \|_F^2  \\
		& +\frac{\eta}{2} \| \mathcal{E} - \mathcal{E}^k \|_F^2+\frac{\mu}{2} \| \mathcal{C}-\mathcal{E} \|_F^2 .
		\label{E_T}
	\end{aligned}
\end{equation}
Following the approach used for subproblem \eqref{L_T}, \eqref{E_T} can be rewritten as:
\begin{equation}
	\min_{\mathcal{E}}  ~  \| \mathcal{E}  \times_1 \mathbf{P}_1 \times_2 \mathbf{P}_2 \times_3 \mathbf{D}_{\mathcal{E}} - \mathcal{Y}_2  \|_F^2  
	+ \xi_2 \| \mathcal{E} - \mathcal{M}_2 \|_F^2,
	\label{E_E}
\end{equation}
where $\mathcal{Y}_2=\mathcal{Y}-\mathcal{L}  \times_1 \mathbf{P}_1 \times_2 \mathbf{P}_2 \times_3 \mathbf{D}_{\mathcal{L}}$, $\displaystyle \xi_2=\frac{\eta+\mu}{2}$, and $\displaystyle \mathcal{M}_2=\frac{\eta \mathcal{E}^k+\mu \mathcal{C}}{\eta+\mu}$. 
This problem is structurally identical to~\eqref{L_L}, and can therefore be solved efficiently using the method proposed in~\cite{Li2018}.

\textbf{b.} The update for $\mathcal{C}$ involves minimizing~\eqref{E 2} with respect to $\mathcal{C}$, i.e.,
\begin{equation}
	\min_{\mathcal{C}} ~ \tau \| \mathcal{L} \times_3 (\mathbf{R}+\bigtriangleup \mathbf{R}) \mathbf{D}_{\mathcal{L}} + \mathcal{C} \times_3 (\mathbf{R}+\bigtriangleup \mathbf{R}) \mathbf{D}_{\mathcal{E}} - \mathcal{Z} \|_F^2
	+ \frac{\mu}{2} \| \mathcal{C} - \mathcal{E} \|_F^2 .
	\label{E_C} 
\end{equation}
Analogous to the subproblem \eqref{L_A}, the optimal solution to \eqref{E_C} is given by:
\begin{equation}
	\mathbf{C}^* = {\mathbf{\Omega}_1}^{-1} ~ \mathbf{\Omega}_2,
	\label{E_C_S}
	\nonumber
\end{equation} 
where
\begin{equation}
	\begin{aligned}
		\mathbf{\Omega}_1 &= 2 \tau \left((\mathbf{R} + \bigtriangleup \mathbf{R}) \mathbf{D}_{\mathcal{E}}\right)^T (\mathbf{R} + \bigtriangleup \mathbf{R}) \mathbf{D}_{\mathcal{E}} + \mu \mathbf{I}, \\
		\mathbf{\Omega}_2 &= 2 \tau \left((\mathbf{R} + \bigtriangleup \mathbf{R}) \mathbf{D}_{\mathcal{E}}\right)^T \left(\mathbf{Z} - (\mathbf{R} + \bigtriangleup \mathbf{R}) \mathbf{D}_{\mathcal{L}} \mathbf{L} \right) + \mu \mathbf{E}.
		\nonumber
	\end{aligned}
\end{equation}

\subsubsection{Optimization with respect to $ \bigtriangleup \mathbf{R} $} 
With $\mathcal{L}$ and $\mathcal{E}$ fixed, the optimization problem \eqref{PAO DR} reduces to the following subproblem:
\begin{equation}
	\min_{\bigtriangleup \mathbf{R}} ~
	\tau \| \mathcal{L} \times_3 (\mathbf{R}+\bigtriangleup \mathbf{R}) \mathbf{D}_{\mathcal{L}} + \mathcal{E} \times_3 (\mathbf{R}+\bigtriangleup \mathbf{R}) \mathbf{D}_{\mathcal{E}} - \mathcal{Z} \|_F^2 +\frac{\eta}{2} \| \bigtriangleup \mathbf{R} - \bigtriangleup \mathbf{R}^k \|_F^2. 
	\label{DR}
\end{equation}
Similarly to subproblems \eqref{L_A} and \eqref{E_C}, the optimal solution to \eqref{DR} can be expressed as:
\begin{equation}
	\bigtriangleup \mathbf{R}^* = \mathbf{\Omega}_2 ~ {\mathbf{\Omega}_1}^{-1},
	\nonumber
\end{equation}
where
\begin{equation}
	\begin{aligned}
		\mathbf{\Omega}_1 &= 2 \tau (\mathbf{D}_{\mathcal{L}} \mathbf{L} + \mathbf{D}_{\mathcal{E}} \mathbf{E}) (\mathbf{D}_{\mathcal{L}} \mathbf{L} + \mathbf{D}_{\mathcal{E}} \mathbf{E})^T + \eta \mathbf{I}, \\
		\mathbf{\Omega}_2 &= 2 \tau ( \mathbf{Z} - \mathbf{R} (\mathbf{D}_{\mathcal{L}} \mathbf{L} + \mathbf{D}_{\mathcal{E}} \mathbf{E})) (\mathbf{D}_{\mathcal{L}} \mathbf{L} + \mathbf{D}_{\mathcal{E}} \mathbf{E} )^T
		+ \eta \bigtriangleup \mathbf{R}^k.
		\nonumber
	\end{aligned}
\end{equation}

\section{CONVERGENCE ANALYSIS}
\label{section 5}
In this section, we present the convergence analysis of Algorithm~\ref{alg:PAO}. 
To facilitate the theoretical proof, we define the function $ g $ as follows:
\begin{equation}
	\begin{aligned}
	h(\mathcal{L}, \mathcal{E},  \Delta \mathbf{R}) = & \|  \mathcal{L}  \times_1 \mathbf{P}_1 \times_2 \mathbf{P}_2 \times_3 \mathbf{D}_{\mathcal{L}} + \mathcal{E}  \times_1 \mathbf{P}_1 \times_2 \mathbf{P}_2 \times_3 \mathbf{D}_{\mathcal{E}} - \mathcal{Y} \|_F^2  \\
	& +  \tau \| \mathcal{L} \times_3 (\mathbf{R}+\bigtriangleup \mathbf{R}) \mathbf{D}_{\mathcal{L}} + \mathcal{E} \times_3 (\mathbf{R}+\bigtriangleup \mathbf{R}) \mathbf{D}_{\mathcal{E}} - \mathcal{Z} \|_F^2.
	\label{g}
	\end{aligned}
\end{equation}
Then, we have:
\begin{equation}
	f(\mathcal{L}, \mathcal{E},  \Delta \mathbf{R}) = h(\mathcal{L}, \mathcal{E},  \Delta \mathbf{R}) + \lambda \phi(\mathcal{L}).
	\label{f}
\end{equation}
We define $ \mathcal{W} := (\mathcal{L}, \mathcal{E},  \Delta \mathbf{R}) $ and $ \mathcal{W}^k := (\mathcal{L}^k, \mathcal{E}^k,  \Delta \mathbf{R}^k) $. 
Next we prove that the function $ f $ satisfies the sufficient descent property.

\begin{lemma}
Let $\{\mathcal{W}^k = (\mathcal{L}^k, \mathcal{E}^k,  \Delta \mathbf{R}^k) \}_{k \in \mathbb{N}}$ be a sequence generated by the PAO algorithm. Then the following conclusions hold:
\begin{equation}
	\begin{aligned}
		(i) & \quad f(\mathcal{W}^{k+1}) + \frac{\eta}{2} \|\mathcal{W}^{k+1} - \mathcal{W}^k\|_F^2 \leq f(\mathcal{W}^k), \\
		(ii) & \quad \sum_{k=0}^{+\infty} \|\mathcal{W}^{k+1} - \mathcal{W}^k\|_F^2 < +\infty, 
		\nonumber
	\end{aligned}
\end{equation}
hence $\lim_{k \to +\infty} \|\mathcal{W}^{k+1} - \mathcal{W}^k\| = 0$.
\label{lemma sufficient descent lemma}
\end{lemma}

\textbf{Proof.} (i) 
Since $\mathcal{L}^{k+1}$, $\mathcal{E}^{k+1}$, and $\Delta \mathbf{R}^{k+1}$ are optimal solutions to subproblems \eqref{PAO L}, \eqref{PAO E}, and \eqref{PAO DR}, respectively.
We can derive the following inequalities:
\begin{equation}
		h(\mathcal{L}^{k+1}, \mathcal{E}^k, \Delta \mathbf{R}^k) + \lambda \phi(\mathcal{L}^{k+1}) + \frac{\eta}{2} \| \mathcal{L}^{k+1} - \mathcal{L}^k \|_F^2 \leq h(\mathcal{L}^k, \mathcal{E}^k, \Delta \mathbf{R}^k) + \lambda \phi(\mathcal{L}^k), 
		\nonumber 
\end{equation}

\begin{equation}
		h(\mathcal{L}^{k+1}, \mathcal{E}^{k+1}, \Delta \mathbf{R}^k) + \lambda \phi(\mathcal{L}^{k+1}) + \frac{\eta}{2} \| \mathcal{E}^{k+1} - \mathcal{E}^k \|_F^2 \leq h(\mathcal{L}^{k+1}, \mathcal{E}^k, \Delta \mathbf{R}^k) + \lambda \phi(\mathcal{L}^{k+1}), 
		\nonumber 
\end{equation}

and
\begin{equation}
		h(\mathcal{L}^{k+1}, \mathcal{E}^{k+1}, \Delta \mathbf{R}^{k+1}) + \lambda \phi(\mathcal{L}^{k+1}) + \frac{\eta}{2} \| \Delta \mathbf{R}^{k+1} - \Delta \mathbf{R}^k \|_F^2  \leq h(\mathcal{L}^{k+1}, \mathcal{E}^{k+1}, \Delta \mathbf{R}^k) + \lambda \phi(\mathcal{L}^{k+1}). 
		\nonumber
\end{equation}

Summing up these the inequalities and simplifying, we can obtain
\begin{equation}
	\begin{aligned}
		& h(\mathcal{L}^{k+1}, \mathcal{E}^{k+1}, \Delta \mathbf{R}^{k+1}) + \lambda \phi(\mathcal{L}^{k+1}) 
		+ \frac{\eta}{2} \Big( \| \mathcal{L}^{k+1} - \mathcal{L}^k \|_F^2 
		+ \| \mathcal{E}^{k+1} - \mathcal{E}^k \|_F^2 \nonumber \\
		& + \| \Delta \mathbf{R}^{k+1} - \Delta \mathbf{R}^k \|_F^2 \Big) 
		\leq h(\mathcal{L}^k, \mathcal{E}^k, \Delta \mathbf{R}^k) + \lambda \phi(\mathcal{L}^k),
	\end{aligned}
\end{equation}
which means
\begin{equation}
	f(\mathcal{W}^{k+1}) + \frac{\eta}{2} \| \mathcal{W}^{k+1} - \mathcal{W}^k \|_F^2 \leq f(\mathcal{W}^k).
	\label{conclusion(1)}
\end{equation}

(ii) Let $ N $ be a positive integer. Summing inequality \eqref{conclusion(1)} over $ k = 0 $ to $ N-1 $, we have
\begin{equation}
	\sum_{k=0}^{N-1} f(\mathcal{W}^{k+1}) + \frac{\eta}{2} \sum_{k=0}^{N-1} \Vert \mathcal{W}^{k+1} -\mathcal{W}^k \Vert_F^2 \leq \sum_{k=0}^{N-1} f(\mathcal{W}^k).
	\nonumber
\end{equation}
Upon further simplification, we arrive at
\begin{equation}
	\sum_{k=0}^{N-1} \| \mathcal{W}^{k+1} - \mathcal{W}^k \|_F^2 \leq \frac{2}{\eta} \big( f(\mathcal{W}^0) - f(\mathcal{W}^N) \big).
	\label{conclusion(2)}
\end{equation}

From conclusion (i), we know that the sequence $\{ f(\mathcal{W}^k) \}_{k \in \mathbb{N}}$ is non-increasing.
And as $f$ is lower-bounded, the sequence $\{ f(\mathcal{W}^k) \}_{k \in \mathbb{N}}$ converges.
In inequality \eqref{conclusion(2)}, we take the limit as $N \to +\infty$, we obtain
$\displaystyle \sum_{k=0}^{+\infty} \Vert \mathcal{W}^{k+1} - \mathcal{W}^k \Vert_F^2 < +\infty$.
Consequently, we derive $\lim_{k \to +\infty} \| \mathcal{W}^{k+1} - \mathcal{W}^k \| = 0$. 
Subsequently, we prove $f(\mathcal{W})$ satisfies relative error condition.

\begin{lemma}
	Assume the sequence $\{\mathcal{W}^k = (\mathcal{L}^k, \mathcal{E}^k,  \Delta \mathbf{R}^k) \}_{k \in \mathbb{N}}$ generated by Algorithm~\ref{alg:PAO} is bounded. For each positive integer $k$, there exists $\bm{d}^{k+1} := (\bm{d}_{\mathcal{L}}, \bm{d}_{\mathcal{E}}, \bm{d}_{\Delta \mathbf{R}}) \in \partial f(\mathcal{W}^{k+1})$ such that
	\begin{equation}
		\Vert \bm{d}^{k+1} \Vert \leq b \Vert \mathcal{W}^{k+1} - \mathcal{W}^k \Vert,
		\nonumber
	\end{equation}
	where $b$ is a positive constant and '$\partial f$' represents the subdifferential of $f$.	
	\label{lemma relative error condition}
\end{lemma}

\textbf{Proof.}
According to the definition of $f$, we can obtain:
\begin{equation}
	\partial_{\mathcal{L}} f(\mathcal{L}, \mathcal{E}, \Delta \mathbf{R}) = \nabla_{\mathcal{L}} h(\mathcal{L}, \mathcal{E}, \Delta \mathbf{R}) + \partial_{\mathcal{L}} \phi(\mathcal{L}),
	\label{2_L_1}
\end{equation}

\begin{equation}
	\nabla_{\mathcal{E}} f(\mathcal{L}, \mathcal{E}, \Delta \mathbf{R}) = \nabla_{\mathcal{E}} h(\mathcal{L}, \mathcal{E}, \Delta \mathbf{R}),
	\label{2_E_1}
\end{equation}

and
\begin{equation}
	\nabla_{\Delta \mathbf{R}} f(\mathcal{L}, \mathcal{E}, \Delta \mathbf{R}) = \nabla_{\Delta \mathbf{R}} h(\mathcal{L}, \mathcal{E}, \Delta \mathbf{R}).
	\label{2_DR_1}
\end{equation}

From the optimality conditions of the subproblems \eqref{PAO L}, \eqref{PAO E} and \eqref{PAO DR}, we get:
\begin{equation}
	0 \in \nabla_{\mathcal{L}} h(\mathcal{L}^{k+1}, \mathcal{E}^k, \Delta \mathbf{R}^k) + \partial_{\mathcal{L}} \phi(\mathcal{L}^{k+1}) + \eta (\mathcal{L}^{k+1} - \mathcal{L}^k),
	\label{2_L_2}
\end{equation}

\begin{equation}
	 \nabla_{\mathcal{E}} h(\mathcal{L}^{k+1}, \mathcal{E}^{k+1}, \Delta \mathbf{R}^k) + \eta (\mathcal{E}^{k+1} - \mathcal{E}^k) = 0,
	\label{2_E_2}
\end{equation}

and
\begin{equation}
	\nabla_{\Delta \mathbf{R}}  h(\mathcal{L}^{k+1}, \mathcal{E}^{k+1}, \Delta \mathbf{R}^{k+1}) + \eta (\Delta \mathbf{R}^{k+1} - \Delta \mathbf{R}^k) = 0.
	\label{2_DR_2}
\end{equation}

Combining \eqref{2_L_1} with \eqref{2_L_2} , we obtain
\begin{equation}
	\begin{aligned}
		0  \in \partial_{\mathcal{L}} f(\mathcal{L}^{k+1}, \mathcal{E}^{k+1}, \Delta \mathbf{R}^{k+1})-\nabla_{\mathcal{L}} h(\mathcal{L}^{k+1}, \mathcal{E}^{k+1}, \Delta \mathbf{R}^{k+1}) +\nabla_{\mathcal{L}} h(\mathcal{L}^{k+1}, \mathcal{E}^k, \Delta \mathbf{R}^k) 
		+  \eta (\mathcal{L}^{k+1} -\mathcal{L}^k) .
		\label{relation_L2}
		\nonumber
	\end{aligned}
\end{equation}
Therefore, it follows that
\begin{equation}
		\nabla_{\mathcal{L}} h(\mathcal{L}^{k+1}, \mathcal{E}^{k+1}, \Delta \mathbf{R}^{k+1}) 
		 -\nabla_{\mathcal{L}} h(\mathcal{L}^{k+1}, \mathcal{E}^k, \Delta \mathbf{R}^k) -  \eta (\mathcal{L}^{k+1} -\mathcal{L}^k) \in \partial_{\mathcal{L}} f(\mathcal{L}^{k+1}, \mathcal{E}^{k+1}, \Delta \mathbf{R}^{k+1}).
		\nonumber
\end{equation}

We define $\bm{d}_{\mathcal{L}} := \nabla_{\mathcal{L}} h(\mathcal{L}^{k+1}, \mathcal{E}^{k+1}, \Delta \mathbf{R}^{k+1}) -\nabla_{\mathcal{L}} h(\mathcal{L}^{k+1}, \mathcal{E}^k, \Delta \mathbf{R}^k) -  \eta (\mathcal{L}^{k+1} -\mathcal{L}^k)$, and then we have $\bm{d}_{\mathcal{L}} \in \partial_{\mathcal{L}} f(\mathcal{L}^{k+1}, \mathcal{E}^{k+1}, \Delta \mathbf{R}^{k+1})$. Consequently, we can get
\begin{equation}
	\begin{aligned}
		\Vert \bm{d}_{\mathcal{L}} \Vert &= \Vert \nabla_{\mathcal{L}} h(\mathcal{L}^{k+1}, \mathcal{E}^{k+1}, \Delta \mathbf{R}^{k+1}) -\nabla_{\mathcal{L}} h(\mathcal{L}^{k+1}, \mathcal{E}^k, \Delta \mathbf{R}^k) -  \eta (\mathcal{L}^{k+1} -\mathcal{L}^k) \Vert \\
		&\leq \Vert \nabla_{\mathcal{L}} h(\mathcal{L}^{k+1}, \mathcal{E}^{k+1}, \Delta \mathbf{R}^{k+1})-\nabla_{\mathcal{L}} h(\mathcal{L}^{k+1}, \mathcal{E}^k, \Delta \mathbf{R}^{k+1}) \Vert \\
		&+ \Vert \nabla_{\mathcal{L}} h(\mathcal{L}^{k+1}, \mathcal{E}^k, \Delta \mathbf{R}^{k+1})-\nabla_{\mathcal{L}} h(\mathcal{L}^{k+1}, \mathcal{E}^k, \Delta \mathbf{R}^k) \Vert + \eta \Vert \mathcal{L}^{k+1} - \mathcal{L}^k \Vert.
		\label{d_L1}
		\nonumber
	\end{aligned}
\end{equation}
From equation \eqref{g}, we can know $h$ is a polynomial function. Since the sequence $\{\mathbf{W}^k\}_{k \in \mathbb{N}}$ is bounded, there exists a compact convex set $\mathbb{W}$ such that $\nabla h$ is Lipschitz continuous in $\mathbb{W}$ with Lipschitz constant $l > 0$. Thus, we have
\begin{equation}
	\begin{aligned}
	\Vert \bm{d}_{\mathcal{L}} \Vert & \leq 
	\eta \Vert \mathcal{L}^{k+1} - \mathcal{L}^k \Vert
	+l  \Vert \mathcal{E}^{k+1} - \mathcal{E}^k \Vert
	+l \Vert \Delta \mathbf{R}^{k+1} - \Delta \mathbf{R}^k \Vert \\
	& \leq (\eta+l) \big( \Vert \mathcal{L}^{k+1} - \mathcal{L}^k \Vert+ \Vert \mathcal{E}^{k+1} - \mathcal{E}^k \Vert+ \Vert \Delta \mathbf{R}^{k+1} - \Delta \mathbf{R}^k \Vert \big)\\
	&=m \Vert \mathcal{W}^{k+1} - \mathcal{W}^k \Vert,
	\label{d_L}
	\end{aligned}
\end{equation}
where $m = \eta+l$.

Let
\begin{equation}
	\begin{aligned}
		&\bm{d}_{\mathcal{E}} = \nabla_{\mathcal{E}} f(\mathcal{L}^{k+1}, \mathcal{E}^{k+1}, \Delta \mathbf{R}^{k+1}), \\
		&\bm{d}_{\Delta \mathbf{R}} = \nabla_{\Delta \mathbf{R}} f(\mathcal{L}^{k+1}, \mathcal{E}^{k+1}, \Delta \mathbf{R}^{k+1}).
		\label{ER}
	\end{aligned}
\end{equation}
Combining \eqref{2_E_1}, \eqref{2_E_2} and \eqref{ER}, we find
\begin{equation}
		\bm{d}_{\mathcal{E}} = \nabla_{\mathcal{E}} h(\mathcal{L}^{k+1}, \mathcal{E}^{k+1}, \Delta \mathbf{R}^{k+1}) -\nabla_{\mathcal{E}} h(\mathcal{L}^{k+1}, \mathcal{E}^{k+1}, \Delta \mathbf{R}^k)- \eta (\mathcal{E}^{k+1} - \mathcal{E}^k).
		\nonumber
\end{equation}
Thus, we immediately obtain
\begin{equation}
	\begin{aligned}
	\Vert \bm{d}_{\mathcal{E}} \Vert &=  \Vert \nabla_{\mathcal{E}} h(\mathcal{L}^{k+1}, \mathcal{E}^{k+1}, \Delta \mathbf{R}^{k+1}) -\nabla_{\mathcal{E}} h(\mathcal{L}^{k+1}, \mathcal{E}^{k+1}, \Delta \mathbf{R}^k)- \eta (\mathcal{E}^{k+1} - \mathcal{E}^k) \Vert
	\label{d_E} \\
	&\leq \Vert \nabla_{\mathcal{E}} h(\mathcal{L}^{k+1}, \mathcal{E}^{k+1}, \Delta \mathbf{R}^{k+1}) -\nabla_{\mathcal{E}} h(\mathcal{L}^{k+1}, \mathcal{E}^{k+1}, \Delta \mathbf{R}^k) \Vert+ \eta \Vert  \mathcal{E}^{k+1} - \mathcal{E}^k \Vert \\
	&\leq l  \Vert \Delta \mathbf{R}^{k+1} - \Delta \mathbf{R}^k  \Vert +\eta \Vert \mathcal{E}^{k+1} - \mathcal{E}^k  \Vert \\
	&\leq m \Vert \mathcal{L}^{k+1} - \mathcal{L}^k \Vert
	+ m  \Vert \mathcal{E}^{k+1} - \mathcal{E}^k \Vert
	+ m \Vert \Delta \mathbf{R}^{k+1} - \Delta \mathbf{R}^k \Vert\\
	&=m \Vert \mathcal{W}^{k+1} - \mathcal{W}^k \Vert.
	\end{aligned}
\end{equation}

Similarly, combining \eqref{2_DR_1}, \eqref{2_DR_2} and \eqref{ER}, we get
\begin{equation}
	\begin{aligned}
		\bm{d}_{\Delta \mathbf{R}} &= \nabla_{\Delta \mathbf{R}} h(\mathcal{L}^{k+1}, \mathcal{E}^{k+1}, \Delta \mathbf{R}^{k+1})-\nabla_{\Delta \mathbf{R}} h(\mathcal{L}^{k+1}, \mathcal{E}^{k+1}, \Delta \mathbf{R}^{k+1}) - \eta (\Delta \mathbf{R}^{k+1} - \Delta \mathbf{R}^k) \\
		&= - \eta (\Delta \mathbf{R}^{k+1} - \Delta \mathbf{R}^k).
		\nonumber
	\end{aligned}
\end{equation}
It follows that
\begin{equation}
	\Vert \bm{d}_{\Delta \mathbf{R}} \Vert = \eta \Vert \Delta \mathbf{R}^{k+1} - \Delta \mathbf{R}^k \Vert \leq \eta \Vert \mathcal{W}^{k+1} - \mathcal{W}^k \Vert.
	\label{d_R}
\end{equation}

From [Proposition 2.1,\cite{Attouch2011}], combining \eqref{d_L}, \eqref{d_E}and \eqref{d_R}, we have
$
\bm{d}^{k+1} := (\bm{d}_{\mathcal{L}}, \bm{d}_{\mathcal{E}}, \bm{d}_{\Delta \mathbf{R}}) \in \partial f(\mathcal{W}^{k+1})
$
and
$
\Vert \bm{d}^{k+1} \Vert \leq \Vert \bm{d}_{\mathcal{L}} \Vert + \Vert \bm{d}_{\mathcal{E}} \Vert +  \Vert \bm{d}_{\Delta \mathbf{R}} \Vert \leq (2 m + \eta) \Vert \mathcal{W}^{k+1} - \mathcal{W}^k \Vert,
$
where $b = 2 m + \eta=2 l + 3 \eta$. 
All in all, the proof is complete.

\begin{lemma}
	The function $f(\mathcal{L}, \mathcal{E}, \Delta \mathbf{R})$ is a KL function.
	\label{lemma KL}
\end{lemma}

\textbf{Proof.} According to \cite{Bochnak1998}, the Frobenius norm is semi-algebraic. Additionally, the regularizer function $\phi(\cdot)$ is also semi-algebraic. Since the function $f(\mathcal{L}, \mathcal{E}, \Delta \mathbf{R})$ can be represented as the sum of a finite number of semi-algebraic functions, it follows that $f(\mathcal{L}, \mathcal{E}, \Delta \mathbf{R})$ is semi-algebraic. Moreover, since $f(\mathcal{L}, \mathcal{E}, \Delta \mathbf{R})$ is proper and continuous, we conclude that $f(\mathcal{L}, \mathcal{E}, \Delta \mathbf{R})$ is a KL function by [Theorem 3,\cite{Bolte2013}].

\begin{theorem}
	Suppose the sequence $\{\mathcal{W}^k = (\mathcal{L}^k, \mathcal{E}^k, \Delta \mathbf{R}^k)\}_{k \in \mathbb{N}}$ generated by Algorithm~\ref{alg:PAO} is bounded. Then the sequence converges to a critical point of $f$ in model \eqref{f} and 
	\begin{equation}
		\sum_{k=0}^{+\infty} \| \mathcal{W}^{k+1} - \mathcal{W}^k \| < +\infty.
		\label{theorem_e}
	\end{equation}
	\label{theorem}
\end{theorem}

\textbf{Proof.} By combining Lemma~\ref{lemma sufficient descent lemma} with Lemma~\ref{lemma relative error condition}, it follows that the $ f $ satisfies both the sufficient descent condition and the relative error condition. Furthermore, since the bounded sequence $\{\mathcal{W}^k\}_{k \in \mathbb{N}}$ admits a convergent subsequence and $f$ is continuous, the continuity condition holds naturally. In addition, Lemma~\ref{lemma KL} establishes that $ f $ is a KL function. Therefore, according to [Theorem 2.9,\cite{Attouch2011}], we conclude that the sequence $\{\mathcal{W}^k\}_{k \in \mathbb{N}}$ generated by Algorithm~\ref{alg:PAO} converges to a critical point of $ f $, and we have 
$
\displaystyle \sum_{k=0}^{+\infty} \Vert \mathcal{W}^{k+1} - \mathcal{W}^k \Vert < +\infty.
$

\begin{remark}[Stopping Criterion]
Based on \eqref{theorem_e} in Theorem~\ref{theorem}, we obtain $\lim_{k \to +\infty} \| \mathcal{W}^{k+1} - \mathcal{W}^k \| = 0$.
Therefore, the relative change between two consecutive steps can be directly employed as the termination criterion for the Algorithm~\ref{alg:PAO}. 
Specifically, we define the relative change at the $k$-th iteration:
\begin{equation}
   \eta_k := \max\left\{ \frac{\|\mathcal{L}^{k+1} - \mathcal{L}^k\|}{\|\mathcal{L}^k\| },\, 
   \frac{\|\mathcal{E}^{k+1} - \mathcal{E}^k\|}{\|\mathcal{E}^k\| },\, 
   \frac{\|\Delta\mathbf{R}^{k+1} - \Delta\mathbf{R}^k\|}{\|\Delta\mathbf{R}^k\| } 
   \right\},
   \label{convergence}
\end{equation}
then the algorithm is terminated when $\eta_k < \varepsilon$, for a prescribed tolerance $\varepsilon > 0$.
\end{remark}

\section{EXPERIMENTAL RESULTS}
\label{section 6}
In this section, we conduct numerous numerical experiments to evaluate the performance of DLRRF model for fusing HSIs and MSIs with inter-image variability. 
In particular, we compare it with five classic fusion methods, including pansharpening method (GLPHS~\cite{Aiazzi2006}), MF-based methods (CNMF~\cite{Yokoya2012} and HySure~\cite{Simoes2014}), and TF-based methods (STEREO~\cite{Kanatsoulis2018a} and SCOTT~\cite{Prevost2020}). Besides, we also compare three methods specifically designed to handle inter-image variability: FuVar~\cite{Borsoi2020}, CB-STAR~\cite{Borsoi2021}, and BTD-Var~\cite{Prevost2022}.
All parameters are meticulously tuned in accordance with the original works. 
Experiments are conducted using MATLAB R2023b on a 13th Gen Intel\textregistered{} Core\texttrademark{} i5-13500H processor running at 2.60 GHz. 

\subsection{Experimental Datasets}
\begin{figure}[htbp]
	\centering
	\begin{subfigure}[t]{0.125\textwidth}
		\centering
		\caption*{HSI}
		\vspace{-10pt}
		\rotatebox{-90}{\includegraphics[height=1.7cm]{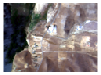}}
		\vspace{-10pt}
	\end{subfigure}\hspace{-15pt}
	\begin{subfigure}[t]{0.125\textwidth}
		\centering
		\caption*{MSI}
		\vspace{-10pt}
		\rotatebox{-90}{\includegraphics[height=1.7cm]{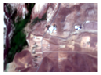}}
		\vspace{-10pt}
	\end{subfigure}\hspace{2pt}
	\begin{subfigure}[t]{0.125\textwidth}
		\centering
		\caption*{HSI}
		\vspace{-3pt}
		\includegraphics[height=2.3cm]{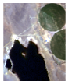}
		\vspace{-10pt}
	\end{subfigure}\hspace{-8pt}
	\begin{subfigure}[t]{0.125\textwidth}
		\centering
		\caption*{MSI}
		\vspace{-3pt}
		\includegraphics[height=2.3cm]{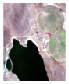}
		\vspace{-10pt}
	\end{subfigure}\hspace{6pt}
	\begin{subfigure}[t]{0.125\textwidth}
		\centering
		\caption*{HSI}
		\vspace{-3pt}
		\includegraphics[height=2.3cm]{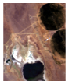}
		\vspace{-10pt}
	\end{subfigure}\hspace{-8pt}
	\begin{subfigure}[t]{0.125\textwidth}
		\centering
		\caption*{MSI}
		\vspace{-3pt}
		\includegraphics[height=2.3cm]{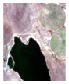}
		\vspace{-10pt}
	\end{subfigure}\hspace{3pt}
	\begin{subfigure}[t]{0.125\textwidth}
		\centering
		\caption*{HSI}
		\vspace{-10pt}
		\rotatebox{-90}{\includegraphics[height=1.5cm]{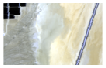}}
		\vspace{-10pt}
	\end{subfigure}\hspace{-18pt}
	\begin{subfigure}[t]{0.125\textwidth}
		\centering
		\caption*{MSI}
		\vspace{-10pt}
		\rotatebox{-90}{\includegraphics[height=1.5cm]{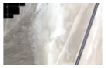}}
		\vspace{-10pt}
	\end{subfigure}
	
	\vspace{1em} 
	
	\makebox[0.2\textwidth][c]{(a)Isabella Lake}\hspace{20pt}%
	\makebox[0.2\textwidth][c]{(b)Lake Tahoe A}\hspace{30pt}%
	\makebox[0.2\textwidth][c]{(c)Lake Tahoe B}\hspace{20pt}%
	\makebox[0.2\textwidth][c]{(d)Ivanpah Playa}
	
	\captionsetup{font=small}
	\caption{Hyperspectral and multispectral images used in the experiments.}
	\label{Fig:HSI_MSI}
\end{figure}

To assess the performance of the DLRRF, we select four HSI and MSI datasets with inter-image variability.
These image sets are categorized into two groups: one pair with a short acquisition interval (less than three months), and three pairs with long acquisition intervals (more than one year). 
The HSIs obtained by the AVIRIS sensor all include $H = 173$ spectral bands, and the MSIs captured by Sentinel-2A all include $h = 10$ spectral bands.
Further details on each dataset are provided as follows:

(1) The first dataset comprises an $80 \times 80$ pixel region over Lake Isabella, captured on 2018-06-27 and 2018-08-27 \cite{Borsoi2020}. 
A true-color visualization of the HSI and MSI is presented in \autoref{Fig:HSI_MSI}(a). 
Despite the short temporal interval, visible discrepancies still exists, particularly in overall tone and specific regions such as the upper-right portion of the lake.

(2) Two distinct image pairs (referred to as Lake Tahoe A and Lake Tahoe B) covering the Lake Tahoe area are used, each of size $100 \times 80$ pixels. 
Lake Tahoe A \cite{Borsoi2020} was recorded on 2014-10-04 and 2017-10-24, while Lake Tahoe B \cite{Wang2023} was collected on 2014-09-19 and 2017-10-24. 
True-color representations are displayed in \autoref{Fig:HSI_MSI}(b) and \autoref{Fig:HSI_MSI}(c), respectively. 
Significant discrepancies between the HSI and MSI are evident in both cases, owing to the long temporal gap. 
For Lake Tahoe A, noticeable changes include shifts in ground color tones and crop patterns. 
Moreover, an island visible in the HSI is absent in the corresponding MSI. 
In Lake Tahoe B, the lake appears significantly larger in the MSI compared to the corresponding HSI.

(3) The last dataset consists of $80 \times 128$ pixel region over Ivanpah Playa, acquired on 2015-10-26 and 2017-12-17 \cite{Borsoi2020}. 
The true-color visualization is showed in \autoref{Fig:HSI_MSI}(d). 
Due to the long acquisition time span, substantial differences can be observed, especially in the central region where variations in sand coloration are clearly noticeable.

\subsection{Experimental Settings} 
In our experimentals, the HR-HSIs and MSIs, acquired at different time instants but with the same spatial resolution, are preprocessed following \cite{Simoes2014}, which includes the removal of water absorption and low signal-to-noise ratio (SNR) bands, followed by normalization of all spectral channels such that the 0.999 intensity quantile equals 1. 
Next, the HR-HSI is denoised using the approach mentioned in \cite{Roger1996}, yielding the reference image $\mathcal{X}$.
For convenience, we use the ground truth (GT) to refer to the reference image $\mathcal{X}$ in the discussion below. 
The observed HSI $\mathcal{Y}$ is derived from $\mathcal{X}$ through a separable degradation model, involving a Gaussian filter with unit variance, followed by downsampling with sampling factors of two or four (described by $sf=2$ and $sf=4$, respectively), and the addition of Gaussian noise to achieve an SNR of $30$ dB. 
Meanwhile, the observed MSI $\mathcal{Z}$ is obtained by adding Gaussian noise into the original MSI to attain an SNR of $40$ dB. 

The spectral response matrix $R$ is derived from calibration measurements. 
Other algorithm parameters are set as follows: 
the dimensionality of the low-rank subspace $S_1$ is typically set within the range of 2 to 6, and the dimensionality of the residual subspace $S_2$ is fixed at 2. 
Furthermore, we set $\tau=1 \times 10^{-1}$, $\lambda=1 \times 10^{-2}$, $\eta=1 \times 10^{-2}$, and $\mu=5 \times 10^{-2}$. 

%
%

\subsection{Evaluation Indexes}
To comprehensively evaluate the quality of the fusion results, denoted as $\widehat{\mathcal{X}}$, we compare them with the reference HR-HSIs, represented by $\mathcal{X}$. 
Specifically, we employ six quantitative evaluation metrics, which are defined in detail below.

The first metric is the \textit{Peak Signal-to-Noise Ratio} (PSNR)~\cite{Borsoi2021}, which is defined as:
\begin{equation}
	\text{PSNR} = \frac{1}{S} \sum_{s=1}^{S} 10 \log_{10} \left( \frac{ W H ~ \mathbb{E} \{ \max \left( [\mathcal{X}]_{:,:,s} \right) \}}{\left\| [\mathcal{X}]_{:,:,s} - [\widehat{\mathcal{X}}]_{:,:,s} \right\|_F^2} \right),
	\nonumber
\end{equation}
where $\mathbb{E} (\cdot)$ denotes the expectation operator. Higher PSNR values indicate better spatial reconstruction quality.

The second metric is the \textit{Structural Similarity} (SSIM)~\cite{Wang2004}, which incorporates human visual perception by assessing errors in correlation, luminance and contrast. SSIM is expressed as:
\begin{equation}
	\text{SSIM} = \frac{(2\mu_{\mathcal{X}}\mu_{\widehat{\mathcal{X}}} + c_1)(2\sigma_{\mathcal{X}\widehat{\mathcal{X}}} + c_2)}{(\mu_{\mathcal{X}}^2 + \mu_{\widehat{\mathcal{X}}}^2 + c_1)(\sigma_{\mathcal{X}}^2 + \sigma_{\widehat{\mathcal{X}}}^2 + c_2)},
	\nonumber
\end{equation}
where $\mu_{\mathcal{X}}$, $\mu_{\widehat{\mathcal{X}}}$, $\sigma_{\mathcal{X}}$, $\sigma_{\widehat{\mathcal{X}}}$, and $\sigma_{\mathcal{X} \widehat{\mathcal{X}}}$ represent local means, standard deviations, and cross-covariance between HSI $\mathcal{X}$ and $\widehat{\mathcal{X}}$, and $c_1$, $c_2$ are small constants to avoid division by zero. The SSIM value ranges between $0$ and $1$, where $0$ indicates no similarity and $1$ indicates perfect reconstruction.

The third metric is the \textit{Erreur Relative Globale Adimensionnelle de Synthèse} (ERGAS)~\cite{Borsoi2021}, which provides a global statistical measure of fused image quality. It is defined as:
\begin{equation}
	\text{ERGAS} = \frac{W H}{w h} \sqrt{ \frac{10^4}{S} \sum_{s=1}^{S} \frac{ \left\| [\mathcal{X}]_{:,:,s} - [\widehat{\mathcal{X}}]_{:,:,s} \right\|_F^2 }{ \left( \frac{1}{W H} \textbf{1}^T [\mathcal{X}]_{:,:,s} \textbf{1} \right)^2 } }.
	\nonumber
\end{equation}
Lower ERGAS value indicates better performance.

The fourth metric is the \textit{Spectral Angle Mapper} (SAM)~\cite{Borsoi2021}, which quantifies the spectral distortion between the reference and fused images. It is calculated as:
\begin{equation}
	\text{SAM} = \frac{1}{W H} \sum_{w,h} \arccos \left( \frac{[\mathcal{X}]_{w,h,:}^T [\widehat{\mathcal{X}}]_{w,h,:}}{\| [\mathcal{X}]_{w,h,:} \|_2 \| [\widehat{\mathcal{X}}]_{w,h,:} \|_2} \right).
	\nonumber
\end{equation}

The fifth metric is the \textit{Root Mean Square Error} (RMSE) \cite{Prevost2022}, which is a standard way to measure the error between the reference and fused images. It is given by:
\begin{equation}
	\text{RMSE} = \sqrt{\frac{1}{W H S} \sum_{w=1}^{W} \sum_{h=1}^{H} \sum_{s=1}^{S} ([\mathcal{X}]_{w,h,s} - [\widehat{\mathcal{X}}]_{w,h,s})^2}.
	\nonumber
\end{equation}

The sixth metric is the \textit{Universal Image Quality Index} (UIQI)~\cite{Wang2002}, which evaluates image distortions including correlation loss, luminance and contrast distortions. The UIQI is defined as:
\begin{equation}
	\text{UIQI} = \frac{4 \sigma_{\mathcal{X},\widehat{\mathcal{X}}} \mu_{\mathcal{X}} \mu_{\widehat{\mathcal{X}}}}{(\sigma_{\mathcal{X}}^2 + \sigma_{\widehat{\mathcal{X}}}^2)(\mu_{\mathcal{X}}^2 + \mu_{\widehat{\mathcal{X}}}^2)}.
	\nonumber
\end{equation}
The UIQI ranges from $-1$ to $1$, with a value of $1$ indicating perfect reconstruction.

Moreover, for visual assessment of the reconstructed images, we present one true-color image in the visible spectrum (assigning the bands at $0.45$, $0.56$, and $0.66$ $\mu m$ to the red, green, and blue channels, respectively) and one pseudo-color image in the infrared spectrum (with the bands at $0.80$, $1.50$, and $2.20$ $\mu m$ assigned to the red, green, and blue channels, respectively).

\subsection{Experimental Results}
\subsubsection{The HSIs and MSIs with a Small Acquisition Time Difference}
Initially, we evaluate the fusion performance of the DLRRF on the dataset with small acquisition time difference. 
Here, the Isabella Lake dataset is used, which exhibits moderate inter-image variability between the HSI and MSI.
\autoref{table:Isabella_Lake} presents the quantitative results, where the best and second-best values are highlighted in bold and underlined, respectively.
When $sf=2$, our method achieves competitive performance, with the SSIM metric ranking second only to FuVar's best result.
Although our method does not achieve a significant improvement over CB-STAR when $sf=4$, DLRRF still delivers the highest PSNR among other comparing methods.
In conclusion, DLRRF exhibits robust and competitive performance across different spatial scaling factors.

\begin{table}[htbp]
	\centering
	\captionsetup{font=small}
	\caption{The experimental results of DLRRF method on the Isabella Lake dataset.}
	\label{table:Isabella_Lake}
	\footnotesize 
	\setlength{\tabcolsep}{3.5pt} 
	\begin{tabular}{l|cccccc|cccccc}
		\hline
		\multirow{2}{*}{Method} 
		& \multicolumn{6}{c|}{$sf=2$} 
		& \multicolumn{6}{c}{$sf=4$} \\
		\cline{2-13}
		& PSNR & SSIM & ERGAS & SAM & RMSE & UIQI 
		& PSNR & SSIM & ERGAS & SAM & RMSE & UIQI \\
		\hline
		GLPHS & 22.819 & 0.504 & 8.442 & 5.114 & 19.191 & 0.822  
		& 18.166 & 0.268 & 7.092 & 7.960 & 32.925 & 0.611 \\
		CNMF & 21.386 & 0.583 & 9.849 & 5.057 & 22.489 & 0.772 
		& 19.956 & 0.532 & 6.090 & 7.035 & 26.226 & 0.667  \\
		HySure & 19.015 & 0.665 & 13.105 & 5.394 & 28.885 & 0.797 
		& 16.553 & 0.605 & 8.643 & 6.301 & 38.645 & 0.684  \\
		STEREO & 21.400 & 0.606 & 25.965 & 27.054 & 58.908 & 0.674 
		& 17.268 & 0.412 & 13.648 & 28.884 & 61.998 & 0.582 \\
		SCOTT & 8.621 & 0.251 & 42.091 & 19.975 & 98.691 & 0.151 
		& 7.855 & 0.177 & 22.763 & 22.506 & 107.289 & 0.068 \\
		FuVar & 29.037 & \textbf{0.893} & 6.691 & 4.184 & 9.337 & \underline{0.959}
		& 24.264 & \textbf{0.745} & 4.886 & 6.436 & 16.004 & 0.896  \\
		CB-STAR & \underline{29.225} & 0.868 & \underline{4.817} & \underline{4.043} & \underline{9.111} & 0.952 
		& \underline{26.074} & \underline{0.723} & \textbf{3.250} & \underline{4.563} & \underline{12.902} & \textbf{0.911} \\
		BTD-Var & 26.117 & 0.751 & 6.821 & 5.604 & 12.744 & 0.907 
		& 22.119 & 0.645 & 4.895 & 7.581 & 20.611 & 0.823  \\
		Ours & \textbf{30.645} & \underline{0.888} & \textbf{4.776} & \textbf{3.015} & \textbf{7.588} & \textbf{0.966} 
		& \textbf{26.176} & 0.694 & \underline{3.798} & \textbf{4.408} & \textbf{12.697} & \underline{0.901 } \\
		\hline
	\end{tabular}
	
	\medskip\noindent\textit{sf=2 and sf=4 represent downsampling factors of two and four, respectively.}
\end{table}

\begin{figure*}[htbp]
	\centering
	
	\captionsetup[subfloat]{labelsep=none,format=plain,labelformat=empty}
	\subfloat[]{\rotatebox{-90}{\includegraphics[height=1.55cm]{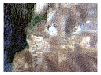}}}\!
	\subfloat[]{\rotatebox{-90}{\includegraphics[height=1.55cm]{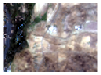}}}\!
	\subfloat[]{\rotatebox{-90}{\includegraphics[height=1.55cm]{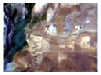}}}\!
	\subfloat[]{\rotatebox{-90}{\includegraphics[height=1.55cm]{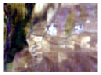}}}\!
	\subfloat[]{\rotatebox{-90}{\includegraphics[height=1.55cm]{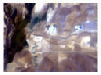}}}\!
	\subfloat[]{\rotatebox{-90}{\includegraphics[height=1.55cm]{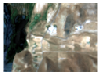}}}\!
	\subfloat[]{\rotatebox{-90}{\includegraphics[height=1.55cm]{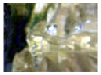}}}\!
	\subfloat[]{\rotatebox{-90}{\includegraphics[height=1.55cm]{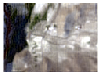}}}\!
	\subfloat[]{\rotatebox{-90}{\includegraphics[height=1.55cm]{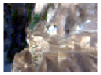}}}\!
	\subfloat[]{\rotatebox{-90}{\includegraphics[height=1.55cm]{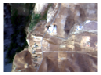}}}
	
	\vspace{-25pt}
	\captionsetup[subfloat]{labelsep=none,format=plain,labelformat=empty}
	\subfloat[GLPHS]{\rotatebox{-90}{\includegraphics[height=1.55cm]{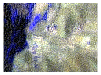}}}\!
	\subfloat[CNMF]{\rotatebox{-90}{\includegraphics[height=1.55cm]{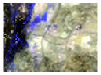}}}\!
	\subfloat[HySure]{\rotatebox{-90}{\includegraphics[height=1.55cm]{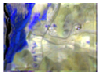}}}\!
	\subfloat[STEREO]{\rotatebox{-90}{\includegraphics[height=1.55cm]{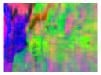}}}\!
	\subfloat[SCOTT]{\rotatebox{-90}{\includegraphics[height=1.55cm]{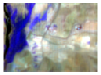}}}\!
	\subfloat[FuVar]{\rotatebox{-90}{\includegraphics[height=1.55cm]{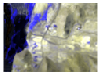}}}\!
	\subfloat[CB-STAR]{\rotatebox{-90}{\includegraphics[height=1.55cm]{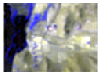}}}\!
	\subfloat[BTD-Var]{\rotatebox{-90}{\includegraphics[height=1.55cm]{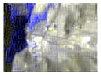}}}\!
	\subfloat[Ours]{\rotatebox{-90}{\includegraphics[height=1.55cm]{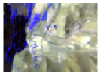}}}\!
	\subfloat[GT]{\rotatebox{-90}{\includegraphics[height=1.55cm]{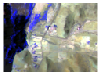}}}

	\captionsetup{font=small}
	\caption{The first and second rows show the visible and infrared representation for the reconstructed images of the Isabella Lake dataset when $sf=2$, respectively.}
	\label{Fig:Isabella_Lake}
\end{figure*}

The visualization results for Isabella Lake ($sf=2$) are presented in \autoref{Fig:Isabella_Lake}. 
We can find our fused images closely resemble the ground truth in both visible and infrared spectral bands. 
The methods encompassing GLPHS, CNMF, HySure, STEREO, and SCOTT all present various levels of blurriness, color distortions, and artifacts within the images, which can be attributed to their inadequate capability in handling inter-image variability. 
Even these approaches considering inter-image variability still present a notable degree of color distortion, such as the visible bands of FuVar and CB-STAR. 
Moreover, BTD-Var produces noticeable blurring and color distortion.

\subsubsection{The HSIs and MSIs With a Large Acquisition Time Difference}
Next, we evaluate the fusion performance of the DLRRF method on the HSI and MSI datasets acquired on large acquisition time difference.
The experiments are conducted on three distinct datasets: Lake Tahoe A, Lake Tahoe B, and Ivanpah Playa.
The quantitative results for each dataset are summarized in \autoref{table:Tahoe_A}, \ref{table:Tahoe_B}, and \ref{table:Ivanpah_Playa}, respectively.
For clarity, the best-performing values are highlighted in bold, and the second-best results are underlined.
The results clearly demonstrate the fusion performance achieved by the proposed DLRRF is superior to that of all other methods.
Particularly, our method achieves significantly higher PSNR values on Lake Tahoe A and Lake Tahoe B datasets compared to all other competing approaches in case of $sf=4$. 
When $sf=2$, despite our method does not achieve a substantial performance improvement over the second-best method, it still outperforms other comparing methods across all metrics.

\begin{table}[htbp]
	\centering
	\captionsetup{font=small}
	\caption{The experimental results of DLRRF method on the Lake Tahoe A dataset.}
	\label{table:Tahoe_A}
	\footnotesize
	\setlength{\tabcolsep}{3.5pt}
	\begin{tabular}{l|cccccc|cccccc}
		\hline
		\multirow{2}{*}{Method} 
		& \multicolumn{6}{c|}{$sf=2$} 
		& \multicolumn{6}{c}{$sf=4$} \\
		\cline{2-13}
		& PSNR & SSIM & ERGAS & SAM & RMSE & UIQI 
		& PSNR & SSIM & ERGAS & SAM & RMSE & UIQI \\
		\hline
		GLPHS & 24.481 & 0.644 & 6.769 & 5.233 & 15.601 & 0.928 
		& 18.451 & 0.302 & 6.669 & 9.695 & 30.894 & 0.771  \\
		CNMF & 21.204 & 0.561 & 9.857 & 5.914 & 22.703 & 0.854  
		& 19.621 & 0.522 & 6.019 & 8.574 & 27.613 & 0.796  \\
		HySure & 20.044 & 0.604 & 11.245 & 8.411 & 26.238 & 0.873 
		& 18.467 & 0.527 & 7.346 & 9.504 & 34.256 & 0.805  \\
		STEREO & 20.407 & 0.617 & 27.623 & 25.378 & 63.446 & 0.692 
		& 17.689 & 0.473 & 14.337 & 27.063 & 65.867 & 0.646 \\
		SCOTT & 10.656 & 0.460 & 38.018 & 31.108 & 88.734 & 0.409 
		& 8.821 & 0.316 & 21.987 & 38.158 & 101.426 & 0.237 \\
		FuVar & 22.148 & 0.802 & 12.204 & 10.135 & 28.128 & 0.904 
		& 23.318 & \underline{0.677} & 3.855 & 6.172 & 18.038 & 0.912  \\
		CB-STAR & \underline{29.751} & \underline{0.878} & \underline{3.655} & \underline{3.515} & \underline{8.505} & \underline{0.979} 
		& \underline{24.073} & 0.637 & \underline{3.499} & 6.713 & \underline{16.256} & \underline{0.927} \\
		BTD-Var & 27.234 & 0.810 & 4.826 & 4.176 & 11.217 & 0.959 
		& 23.168 & 0.647 & 3.880 & \underline{5.547} & 17.924 & 0.909  \\
		Ours & \textbf{30.391} & \textbf{0.909} & \textbf{3.372} & \textbf{2.473} & \textbf{7.806} & \textbf{0.983} 
		& \textbf{25.612} & \textbf{0.746} & \textbf{2.986} & \textbf{4.399} & \textbf{13.848} & \textbf{0.946}  \\
		\hline
	\end{tabular}
\end{table}

\begin{table}[htbp]
	\centering
	\captionsetup{font=small}
	\caption{The experimental results of DLRRF method on the Lake Tahoe B dataset.}
	\label{table:Tahoe_B}
	\footnotesize
	\setlength{\tabcolsep}{3.5pt}
	\begin{tabular}{l|cccccc|cccccc}
		\hline
		\multirow{2}{*}{Method} 
		& \multicolumn{6}{c|}{$sf=2$} 
		& \multicolumn{6}{c}{$sf=4$} \\
		\cline{2-13}
		& PSNR & SSIM & ERGAS & SAM & RMSE & UIQI 
		& PSNR & SSIM & ERGAS & SAM & RMSE & UIQI  \\
		\hline
		GLPHS & 24.420 & 0.525 & 5.487 & 3.572 & 16.052 & 0.856 
		& 19.959 & 0.287 & 4.559 & 6.279 & 26.698 & 0.693  \\
		CNMF & 15.168 & 0.221 & 15.951 & 9.416 & 47.275 & 0.360 
		& 13.433 & 0.291 & 9.460 & 9.017 & 55.427 & 0.232  \\
		HySure & 16.265 & 0.543 & 14.534 & 7.389 & 40.265 & 0.520 
		& 11.825 & 0.376 & 11.447 & 13.365 & 66.413 & 0.257 \\
		STEREO & 19.665 & 0.518 & 25.625 & 27.625 & 73.228 & 0.629 
		& 15.601 & 0.309 & 13.420 & 28.766 & 76.801 & 0.506  \\
		SCOTT & 10.644 & 0.372 & 34.886 & 36.768 & 100.944 & 0.307 
		& 7.513 & 0.223 & 19.771 & 35.911 & 115.123 & 0.146 \\
		FuVar & 27.275 & 0.845 & 4.312 & 3.718 & 11.877 & 0.934 
		& \underline{23.243} & \underline{0.617} & \underline{3.142} & \underline{4.252} & \underline{17.901} & \underline{0.835}  \\
		CB-STAR & \underline{30.890} & \underline{0.875} & \underline{2.632} & \underline{2.159} & \underline{7.650} & \underline{0.967} 
		& 22.839 & 0.513 & 3.331 & 5.388 & 19.675 & 0.814 \\
		BTD-Var & 25.732 & 0.748 & 5.147 & 6.164 & 14.127 & 0.896 
		& 22.067 & 0.539 & 3.582 & 6.996 & 20.410 & 0.809 \\
		Ours & \textbf{31.122} & \textbf{0.888} & \textbf{2.527} & \textbf{1.757} & \textbf{7.350} & \textbf{0.969} 
		& \textbf{25.996} & \textbf{0.719} & \textbf{2.276} & \textbf{2.704} & \textbf{13.317} & \textbf{0.913} \\
		\hline
	\end{tabular}
\end{table}

\begin{table}[htbp]
	\centering
	\captionsetup{font=small}
	\caption{The experimental results of DLRRF method on the Ivanpah Playa dataset.}
	\label{table:Ivanpah_Playa}
	\footnotesize
	\setlength{\tabcolsep}{3.5pt}
	\begin{tabular}{l|cccccc|cccccc}
		\hline
		\multirow{2}{*}{Method} 
		& \multicolumn{6}{c|}{$sf=2$} 
		& \multicolumn{6}{c}{$sf=4$} \\
		\cline{2-13}
		& PSNR & SSIM & ERGAS & SAM & RMSE & UIQI 
		& PSNR & SSIM & ERGAS & SAM & RMSE & UIQI  \\
		\hline
		GLPHS & 27.155 & 0.638 & 3.011 & 1.548 & 11.425 & 0.815 
		& 20.141 & 0.263 & 3.356 & 3.225 & 25.349 & 0.479  \\
		CNMF & 26.637 & 0.690 & 3.162 & 1.152 & 11.923 & 0.776 
		& 23.403 & 0.683 & 2.305 & 1.823 & 17.284 & 0.554  \\
		HySure & 23.648 & 0.729 & 4.512 & 1.945 & 17.015 & 0.609 
		& 21.839 & 0.726 & 2.773 & 2.255 & 20.844 & 0.522 \\
		STEREO & 20.320 & 0.636 & 24.148 & 27.656 & 90.476 & 0.611 
		& 18.041 & 0.542 & 12.371 & 28.372 & 92.791 & 0.502  \\
		SCOTT & 11.582 & 0.447 & 28.323 & 32.160 & 106.644 & 0.423 
		& 7.643 & 0.198 & 17.054 & 38.916 & 128.037 & 0.167  \\
		FuVar & \underline{31.502} & \underline{0.917} & 1.901 & 1.616 & 7.183 & 0.934 
		& \underline{29.029} & \underline{0.858} & \underline{1.240} & 1.773 & \underline{9.327} & \underline{0.903}  \\
		CB-STAR & 31.393 & 0.903 & \underline{1.842} & \underline{1.231} & \underline{6.940} & \underline{0.949} 
		& 27.119 & 0.787 & 1.516 & \underline{1.670} & 11.354 & 0.886  \\
		BTD-Var & 28.202 & 0.809 & 2.745 & 2.004 & 10.428 & 0.884 
		& 25.744 & 0.721 & 1.773 & 2.294 & 13.428 & 0.820  \\
		Ours & \textbf{32.629} & \textbf{0.922} & \textbf{1.606} & \textbf{0.904} & \textbf{6.055} & \textbf{0.957} 
		& \textbf{29.231} & \textbf{0.860} & \textbf{1.196} & \textbf{1.418} & \textbf{9.002} & \textbf{0.924} \\
		\hline
	\end{tabular}
\end{table}

\begin{figure*}[htbp]
	\centering

	\captionsetup[subfloat]{labelsep=none,format=plain,labelformat=empty}
	\subfloat[]{\includegraphics[width=1.55cm]{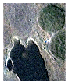}}\!
	\subfloat[]{\includegraphics[width=1.55cm]{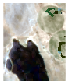}}\!
	\subfloat[]{\includegraphics[width=1.55cm]{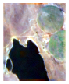}}\!
	\subfloat[]{\includegraphics[width=1.55cm]{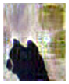}}\!
	\subfloat[]{\includegraphics[width=1.55cm]{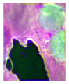}}\!
	\subfloat[]{\includegraphics[width=1.55cm]{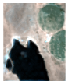}}\!
	\subfloat[]{\includegraphics[width=1.55cm]{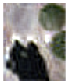}}\!
	\subfloat[]{\includegraphics[width=1.55cm]{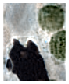}}\!
	\subfloat[]{\includegraphics[width=1.55cm]{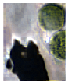}}\!
	\subfloat[]{\includegraphics[width=1.55cm]{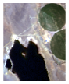}}
	
	\vspace{-25pt}
	\captionsetup[subfloat]{labelsep=none,format=plain,labelformat=empty}
	\subfloat[GLPHS]{\includegraphics[width=1.55cm]{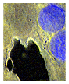}}\!
	\subfloat[CNMF]{\includegraphics[width=1.55cm]{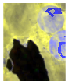}}\!
	\subfloat[HySure]{\includegraphics[width=1.55cm]{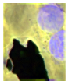}}\!
	\subfloat[STEREO]{\includegraphics[width=1.55cm]{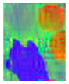}}\!
	\subfloat[SCOTT]{\includegraphics[width=1.55cm]{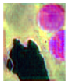}}\!
	\subfloat[FuVar]{\includegraphics[width=1.55cm]{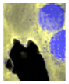}}\!
	\subfloat[CB-STAR]{\includegraphics[width=1.55cm]{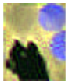}}\!
	\subfloat[BTD-Var]{\includegraphics[width=1.55cm]{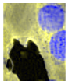}}\!
	\subfloat[Ours]{\includegraphics[width=1.55cm]{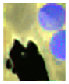}}\!
	\subfloat[GT]{\includegraphics[width=1.55cm]{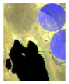}}
	
	\captionsetup{font=small}
	\caption{The first and second rows show the visible and infrared representation for the reconstructed images of the Lake Tahoe A dataset when $sf=4$, respectively.}
	\label{Fig:Tahoe_A}
\end{figure*}

\begin{figure*}[t]
	\centering
	
	\captionsetup[subfloat]{labelsep=none,format=plain,labelformat=empty}
	\subfloat[]{\includegraphics[width=1.55cm]{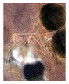}}\!
	\subfloat[]{\includegraphics[width=1.55cm]{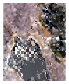}}\!
	\subfloat[]{\includegraphics[width=1.55cm]{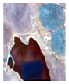}}\!
	\subfloat[]{\includegraphics[width=1.55cm]{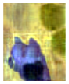}}\!
	\subfloat[]{\includegraphics[width=1.55cm]{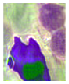}}\!
	\subfloat[]{\includegraphics[width=1.55cm]{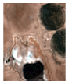}}\!
	\subfloat[]{\includegraphics[width=1.55cm]{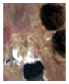}}\!
	\subfloat[]{\includegraphics[width=1.55cm]{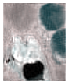}}\!
	\subfloat[]{\includegraphics[width=1.55cm]{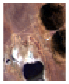}}\!
	\subfloat[]{\includegraphics[width=1.55cm]{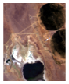}}
	
	\vspace{-25pt}
	\captionsetup[subfloat]{labelsep=none,format=plain,labelformat=empty}
	\subfloat[GLPHS]{\includegraphics[width=1.55cm]{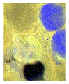}}\!
	\subfloat[CNMF]{\includegraphics[width=1.55cm]{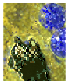}}\!
	\subfloat[HySure]{\includegraphics[width=1.55cm]{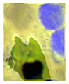}}\!
	\subfloat[STEREO]{\includegraphics[width=1.55cm]{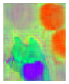}}\!
	\subfloat[SCOTT]{\includegraphics[width=1.55cm]{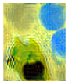}}\!
	\subfloat[FuVar]{\includegraphics[width=1.55cm]{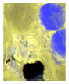}}\!
	\subfloat[CB-STAR]{\includegraphics[width=1.55cm]{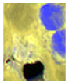}}\!
	\subfloat[BTD-Var]{\includegraphics[width=1.55cm]{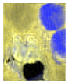}}\!
	\subfloat[Ours]{\includegraphics[width=1.55cm]{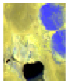}}\!
	\subfloat[GT]{\includegraphics[width=1.55cm]{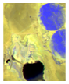}}

	\captionsetup{font=small}
	\caption{The first and second rows show the visible and infrared representation for the reconstructed images of the Lake Tahoe B dataset when $sf=2$, respectively.}
	\label{Fig:Tahoe_B}
\end{figure*}

\begin{figure*}[t]
	\centering
	\captionsetup[subfloat]{labelsep=none,format=plain,labelformat=empty}
	\subfloat[]{\rotatebox{-90}{\includegraphics[height=1.55cm]{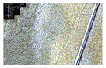}}}\!
	\subfloat[]{\rotatebox{-90}{\includegraphics[height=1.55cm]{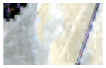}}}\!
	\subfloat[]{\rotatebox{-90}{\includegraphics[height=1.55cm]{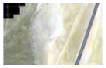}}}\!
	\subfloat[]{\rotatebox{-90}{\includegraphics[height=1.55cm]{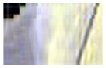}}}\!
	\subfloat[]{\rotatebox{-90}{\includegraphics[height=1.55cm]{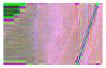}}}\!
	\subfloat[]{\rotatebox{-90}{\includegraphics[height=1.55cm]{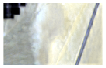}}}\!
	\subfloat[]{\rotatebox{-90}{\includegraphics[height=1.55cm]{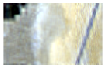}}}\!
	\subfloat[]{\rotatebox{-90}{\includegraphics[height=1.55cm]{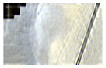}}}\!
	\subfloat[]{\rotatebox{-90}{\includegraphics[height=1.55cm]{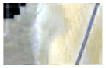}}}\!
	\subfloat[]{\rotatebox{-90}{\includegraphics[height=1.55cm]{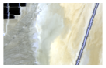}}}
	
	\vspace{-25pt}
	\captionsetup[subfloat]{labelsep=none,format=plain,labelformat=empty}
	\subfloat[GLPHS]{\rotatebox{-90}{\includegraphics[height=1.55cm]{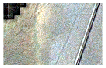}}}\!
	\subfloat[CNMF]{\rotatebox{-90}{\includegraphics[height=1.55cm]{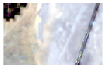}}}\!
	\subfloat[HySure]{\rotatebox{-90}{\includegraphics[height=1.55cm]{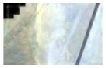}}}\!
	\subfloat[STEREO]{\rotatebox{-90}{\includegraphics[height=1.55cm]{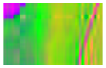}}}\!
	\subfloat[SCOTT]{\rotatebox{-90}{\includegraphics[height=1.55cm]{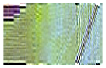}}}\!
	\subfloat[FuVar]{\rotatebox{-90}{\includegraphics[height=1.55cm]{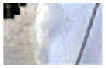}}}\!
	\subfloat[CB-STAR]{\rotatebox{-90}{\includegraphics[height=1.55cm]{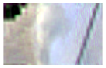}}}\!
	\subfloat[BTD-Var]{\rotatebox{-90}{\includegraphics[height=1.55cm]{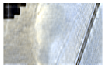}}}\!
	\subfloat[Ours]{\rotatebox{-90}{\includegraphics[height=1.55cm]{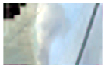}}}\!
	\subfloat[GT]{\rotatebox{-90}{\includegraphics[height=1.55cm]{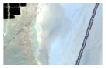}}}
	
	\captionsetup{font=small}
	\caption{The first and second rows show the visible and infrared representation for the reconstructed images of the Ivanpah Playa dataset when $sf=4$, respectively.}
	\label{Fig:Ivanpah_Playa}
\end{figure*}

The visualization results of the reconstructed images are shown in \autoref{Fig:Tahoe_A}, \ref{Fig:Tahoe_B}, and \ref{Fig:Ivanpah_Playa}, providing a comparison of the fusion performance across different methods for the Lake Tahoe A ($sf=4$), Lake Tahoe B ($sf=2$), and Ivanpah Playa ($sf=4$) datasets, respectively. 
From the figures, GLPHS, CNMF, HySure, STEREO and SCOTT exhibit visible artifacts, color distortions or over-sharpening, which can result largely from the fact that these methods fail to consider inter-image variability. 
Among these approaches designed to account for inter-image variability, BTD-Var introduces noticeable artifacts in Lake Tahoe A, severe color distortion in Lake Tahoe B, and blurred stripes in Ivanpah Playa.  
Although FuVar, CB-STAR and our method appear visually similar for Lake Tahoe B in case of $sf=2$, differences become apparent for Lake Tahoe A and Ivanpah Playa datasets in case of $sf=4$. 
Specifically, FuVar and CB-STAR exhibit clear artifacts and substantial color distortion.
In summary, DLRRF achieves high consistency with the reference data in both visible and infrared spectral bands.

\vspace{10pt} 

\begin{figure*}[htbp]
	\centering
	\subfloat[Lake Tahoe A ($sf=2$)]{\includegraphics[height=5cm]{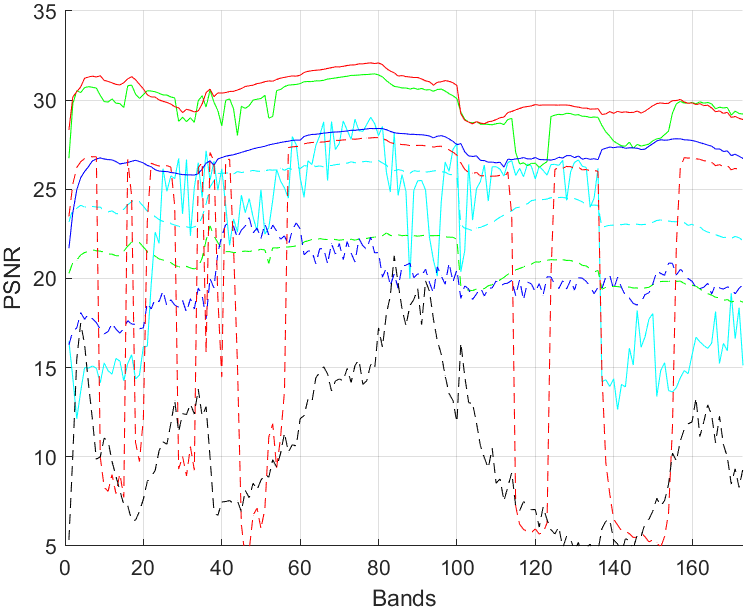}}\hspace{3mm}%
	\subfloat[Lake Tahoe A ($sf=4$)]{\includegraphics[height=5cm]{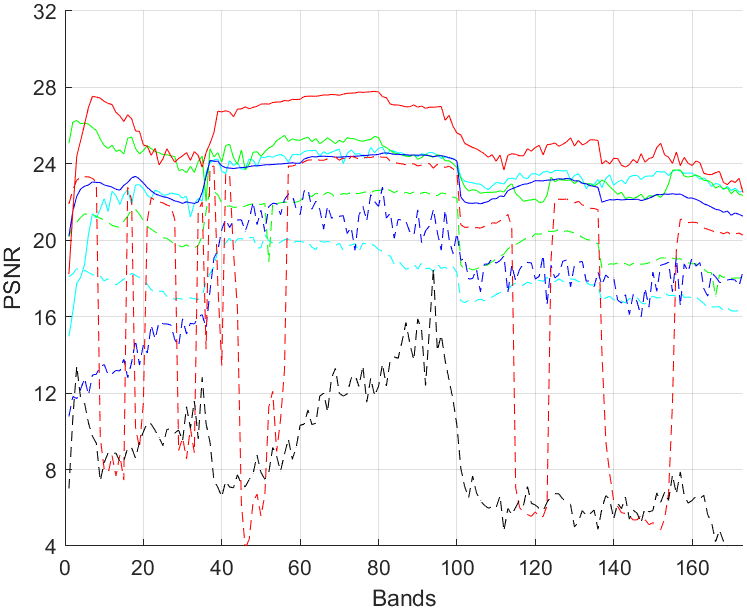}} \\
	
	\subfloat[Lake Tahoe B ($sf=2$)]{\includegraphics[height=5cm]{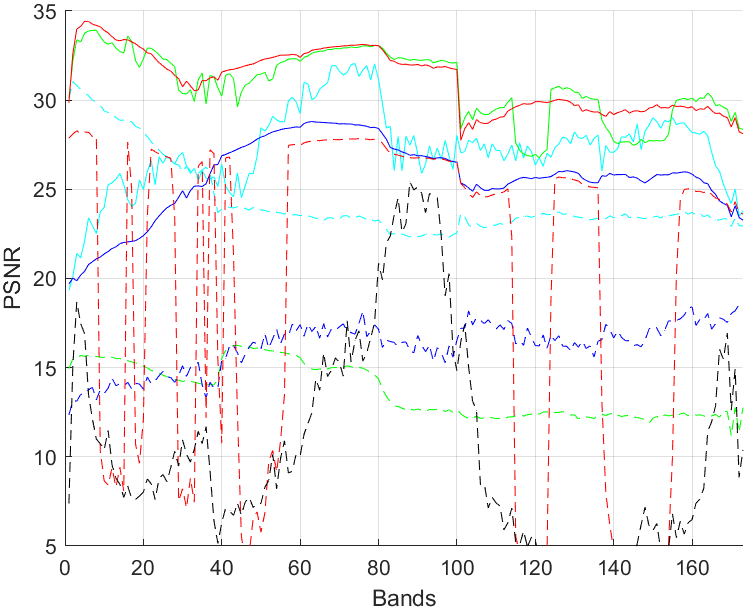}}\hspace{3mm}%
	\subfloat[Lake Tahoe B ($sf=4$)]{\includegraphics[height=5cm]{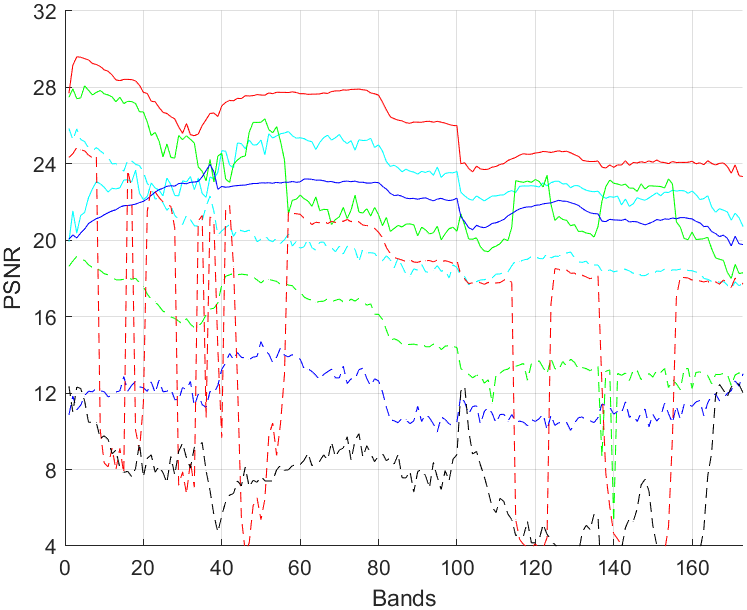}} \\
	
	\vspace{2mm}
	\subfloat{\includegraphics[width=0.6\linewidth]{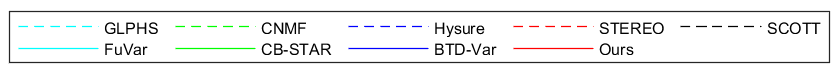}} 
	
	\captionsetup{font=small}
	\caption{The PSNR of different bands of the reconstructed HR-HSIs obtained by compared methods.}
	\label{PSNR}
\end{figure*}

In \autoref{PSNR}, we present the PSNR distributions across all spectral bands for the fused images obtained from the Lake Tahoe A and Lake Tahoe B datasets, with results separately shown under $sf=2$ and $sf=4$. 
As observed in \autoref{PSNR}(b) and (d), the proposed method achieves significantly higher PSNR values across nearly all spectral bands compared to all competing methods under $sf=4$. 
From \autoref{PSNR}(a) and (c), we find that the DLRRF shows only marginal improvement over the second-best method in case of $sf=2$.
Overall, our method attains the highest PSNR in the vast majority of bands under both $sf=2$ and $sf=4$, demonstrating consistent superiority across different scaling factors.
These experiments further highlight the effectiveness of DLRRF in preserving both spectral fidelity and spatial detail during fusion process.

\subsection{Running Time}
Previous experiments show that methods accounting for inter-image variability generally outperform classic fusion approaches, so we focus on comparing their running times in the following analysis. 
Table~\ref{tab:time} presents the running time and corresponding PSNR values of these comparing methods in case of $sf=2$ and $sf=4$, with the best results highlighted in bold and the second-best results underlined. 
From the table, we find that BTD-Var achieves the shortest runtime across all datasets among these methods.
However, it yields significantly inferior fusion performance compared to FuVar, CB-STAR, and DLRRF from the previous experiments.  
Despite the fusion performance of FuVar and CB-STAR are comparable to DLRRF in certain scenarios, the running time of FuVar and CB-STAR is significantly higher than ours. 
In conclusion, our method can strike a favorable balance between computational efficiency and fusion performance.

\begin{table}[htpb]
	\centering
	\captionsetup{font=small}
	\caption{Running time and corresponding PSNR values for different methods, denoted as (TIME(s), PSNR(dB)).}
	\label{tab:time}
	\footnotesize
	\setlength{\tabcolsep}{3pt} 
	\resizebox{\textwidth}{!}{
		\begin{tabular}{l|cccc|cccc}
			\hline
			\multirow{2}{*}{Methods} 
			& \multicolumn{4}{c|}{$sf=2$} 
			& \multicolumn{4}{c}{$sf=4$} \\
			\cline{2-9}
			& Isabella Lake & Lake Tahoe A & Lake Tahoe B & Ivanpah Playa 
			& Isabella Lake & Lake Tahoe A & Lake Tahoe B & Ivanpah Playa \\
			\hline
			FuVar    & (240.6, 29.04) & (110.5, 22.15) & (149.4, 27.28) & (146.8, \underline{31.50}) 
			& (242.8, 24.26) & (242.7, 23.32) & (268.3, \underline{23.24}) & (186.6, \underline{29.03}) \\
			CB-STAR  & (23.0, \underline{29.23}) & (85.3, \underline{29.75}) & (47.1, \underline{30.89}) & (59.5, 31.39) 
			& (55.5, \underline{26.07}) & (40.8, \underline{24.07}) & (41.9, 22.84) & (13.2, 27.12) \\
			BTD-Var  & (\textbf{2.5}, 26.12)  & (\textbf{1.6}, 27.23)  & (\textbf{1.6}, 25.73)  & (\textbf{2.3}, 28.20) 
			& (\textbf{0.9}, 22.12)  & (\textbf{0.8}, 23.17)  & (\textbf{0.9}, 22.07)  & (\textbf{1.1}, 25.74) \\
			Ours     & (\underline{16.2}, \textbf{30.65}) & (\underline{22.1}, \textbf{30.39}) & (\underline{19.6}, \textbf{31.12}) & (\underline{12.7}, \textbf{32.63}) 
			& (\underline{16.7}, \textbf{26.18}) & (\underline{23.7}, \textbf{25.61}) & (\underline{19.6}, \textbf{26.00}) & (\underline{9.0}, \textbf{29.23}) \\
			\hline
		\end{tabular}%
	}
\end{table}

\subsection{Ablation Study}
To assess the contribution of each term in DLRRF model, we perform an ablation study based on the PSNR values of the reconstructed HR-HSIs. 
Using $sf=2$ as a representative case, we examine the roles of the spectral deviation term $\Delta \mathbf{R}$, the residual component $\mathcal{E}$, and the regularizer $\phi(\cdot)$. 
The experiment result is summarized in \autoref{AS PSNR 2}, with the best and second-best PSNR values highlighted in bold and underlined, respectively.

\begin{table}[htbp]
	\centering
	\captionsetup{font=small}
	\caption{Effectiveness evaluation of the $\Delta \mathbf{R}$, $\mathcal{E}$, and $\phi(\mathcal{L})$ terms using PSNR.}
	\label{AS PSNR 2}
	\begin{tabular}{l c c c c}
		\hline
		\multirow{2}{*}{Methods} & \multicolumn{1}{c}{Images with a small acquisition time difference} & \multicolumn{3}{c}{Images with a large acquisition time difference} \\
		\cmidrule(lr){2-2} \cmidrule(lr){3-5}
		& Isabella Lake & Lake Tahoe A & Lake Tahoe B & Ivanpah Playa \\
		\hline
		Without $\Delta \mathbf{R}$   & 30.05 & 29.75 & 27.93 & \underline{32.96} \\
		Without $\mathcal{E}$         & \underline{30.49} & \underline{30.37} & \underline{30.61} & 32.68 \\
		Without $\phi(\mathcal{L})$   & 28.51 & 26.67 & 24.99 & 31.99 \\
		Ours                          & \textbf{30.65} & \textbf{30.93} & \textbf{30.92} & \textbf{33.11} \\
		\hline
	\end{tabular}
\end{table}

Comparing the first and fourth rows, we can find that our model delivers the best reconstruction performance across all datasets, confirming the importance of $\Delta \mathbf{R}$ in modeling spectral variability.
The results of the second and fourth rows reveals that the residual component $\mathcal{E}$ contributes positively to the recovery of lost spatial details.
Furthermore, the contrast between the third and fourth rows presents the effectiveness of the implicit regularization in estimating the tensor $\mathcal{L}$.

\subsection{The Influence of Fusion on Classification Accuracy}
In this section, we conduct classification experiment using the Houston dataset \cite{Debes2014} to evaluate the performance of the DLRRF method in practical applications. This reference HR-HSI contains $349 \times 1905$ spatial pixels and $144$ spectral channels. 
After appropriate preprocessing, we extract spectral bands ranging from $11$ to $110$ and group every $25$ bands by averaging to form the original MSI.
A separable degradation model is employed to generate the observed HSI, involving applying Gaussian filter with standard deviation one and downsampling operation with a decimation factor of four.
Here, we focus only on spectral variability. Specifically, the intensity of the original MSI is appropriately adjusted to obtain the observed MSI.
The DLRRF model fuses observed HSI with MSI to reconstruct the target HSI.
Next, the SVM-KC classifier \cite{CampsValls2006} is applied to classify the HSI, MSI, and target HSI.
The classification results are reported in \autoref{tab:classification}, including per-category accuracy and overall accuracy, where the best results are highlighted in bold.
Most categories achieve improved classification accuracy when the target HSI is used. 
In particular, the overall accuracy of the fused image is 16.8\% higher than that of the HSI and 4.2\% higher than that of the MSI, which reveals the potential of our method for improving classification performance.

\begin{table}[htbp]
	\centering
	\captionsetup{font=small}
	\caption{Classification results of HSI, MSI and HR-HSI (Accuracy(\%)).}
	\label{tab:classification}
	\setlength{\tabcolsep}{1.02pt} 
	\begin{tabular}{l*{16}{c}}
		\hline
		\shortstack[t]{Category} 
		& \shortstack[t]{Healthy\\grass} 
		& \shortstack[t]{Stressed\\grass} 
		& \shortstack[t]{Synthetic\\grass} 
		& \shortstack[t]{Trees} 
		& \shortstack[t]{Soil} 
		& \shortstack[t]{Water} 
		& \shortstack[t]{Residential} 
		& \shortstack[t]{Commercial} 
		& \shortstack[t]{Road} 
		& \shortstack[t]{Highway} 
		& \shortstack[t]{Railway} 
		& \shortstack[t]{Parking\\lot 1} 
		& \shortstack[t]{Parking\\lot 2} 
		& \shortstack[t]{Tennis\\court} 
		& \shortstack[t]{Running\\track} 
		& \shortstack[t]{Average} \\
    	\hline
		HSI & 70 & 67 & 71 & 60 & 76 & 72 & 54 & 52 & 56 & 70 & 54 & \textbf{58} & 62 & \textbf{94} & 87 & 66.9 \\
		MSI & 91 & \textbf{97} & 76 & 94 & 85 & \textbf{95} & \textbf{57} & 58 & 63 & 64 & \textbf{60} & 55 & 45 & 90 & \textbf{95} & 75.0 \\
		HR-HSI & \textbf{92} & \textbf{97} & \textbf{80} & \textbf{95} & \textbf{93} & \textbf{95} & \textbf{57} & \textbf{60} & \textbf{69} & \textbf{73} & 54 & 56 & \textbf{63} & \textbf{94} & 94 & \textbf{78.1}\\
		\hline
	\end{tabular}
\end{table}


\section{CONCLUSION}
\label{section 7}
In this paper, we introduce a new approach for fusing HSIs and MSIs with inter-image variability. 
The core idea lies in modeling spectral variability as change in the spectral degradation operator.
Moreover, we decompose the target HSI into low-rank and residual components, where the latter is used to capture the lost spatial details caused by spatially localized changes.
Leveraging the high spectral correlation of HR-HSI, we project the two components into low-dimension subspaces to reduce computational complexity.
Additionally, we introduce implicit regularization as spatial prior to fully exploit the spatial structural information of the images.
The proposed DLRRF model is solved using the PAO algorithm embedded within the PnP framework, with subproblem involving regularization addressed by an external denoiser. 
We further provide a theoretical convergence analysis of the algorithm.
Finally, extensive experiments evaluate our method against some well-known fusion methods, consistently demonstrating its superior reconstruction performance in the presence of inter-image variability.


\bibliographystyle{unsrt}
\bibliography{Reference}

@Article{Yokoya2017,
  author  = {Naoto Yokoya and Claas Grohnfeldt and Jocelyn Chanussot},
  journal = {IEEE Geoscience and Remote Sensing Magazine},
  title   = {Hyperspectral and Multispectral Data Fusion: A comparative review of the recent literature},
  year    = {2017},
  pages   = {29-56},
  volume  = {5},
  url     = {https://api.semanticscholar.org/CorpusID:28018831},
}

@Article{Prevost2020,
  author   = {Prévost, Clémence and Usevich, Konstantin and Comon, Pierre and Brie, David},
  journal  = {IEEE Transactions on Signal Processing},
  title    = {Hyperspectral Super-Resolution With Coupled Tucker Approximation: Recoverability and SVD-Based Algorithms},
  year     = {2020},
  pages    = {931-946},
  volume   = {68},
  doi      = {10.1109/TSP.2020.2965305},
}

@Article{Li2018,
  author   = {Li, Shutao and Dian, Renwei and Fang, Leyuan and Bioucas-Dias, José M.},
  journal  = {IEEE Transactions on Image Processing},
  title    = {Fusing Hyperspectral and Multispectral Images via Coupled Sparse Tensor Factorization},
  year     = {2018},
  number   = {8},
  pages    = {4118-4130},
  volume   = {27},
  doi      = {10.1109/TIP.2018.2836307},
}

@InProceedings{Zhang2019,
  author    = {Zhang, Guoyong and Fu, Xiao and Huang, Kejun and Wang, Jun},
  booktitle = {2019 IEEE 8th International Workshop on Computational Advances in Multi-Sensor Adaptive Processing (CAMSAP)},
  title     = {Hyperspectral Super-Resolution: A Coupled Nonnegative Block-Term Tensor Decomposition Approach},
  year      = {2019},
  pages     = {470-474},
  doi       = {10.1109/CAMSAP45676.2019.9022476},
}

@Article{Li2022,
  author   = {Jiaxin Li and Danfeng Hong and Lianru Gao and Jing Yao and Ke Zheng and Bing Zhang and Jocelyn Chanussot},
  journal  = {International Journal of Applied Earth Observation and Geoinformation},
  title    = {Deep learning in multimodal remote sensing data fusion: A comprehensive review},
  year     = {2022},
  issn     = {1569-8432},
  pages    = {102926},
  volume   = {112},
  doi      = {https://doi.org/10.1016/j.jag.2022.102926},
  url      = {https://www.sciencedirect.com/science/article/pii/S1569843222001248},
}

@InProceedings{Yao2020,
  author    = {Yao, Jing and Hong, Danfeng and Chanussot, Jocelyn and Meng, Deyu and Zhu, Xiaoxiang and Xu, Zongben},
  booktitle = {European Conference on Computer Vision},
  title     = {Cross-Attention in Coupled Unmixing Nets for Unsupervised Hyperspectral Super-Resolution},
  year      = {2020},
  pages     = {208--224},
  isbn      = {978-3-030-58526-6},
}

@Article{Palsson2017,
  author   = {Palsson, Frosti and Sveinsson, Johannes R. and Ulfarsson, Magnus O.},
  journal  = {IEEE Geoscience and Remote Sensing Letters},
  title    = {Multispectral and Hyperspectral Image Fusion Using a 3-D-Convolutional Neural Network},
  year     = {2017},
  number   = {5},
  pages    = {639-643},
  volume   = {14},
  doi      = {10.1109/LGRS.2017.2668299},
}

@Article{Borsoi2020,
  author   = {Borsoi, Ricardo Augusto and Imbiriba, Tales and Bermudez, José Carlos Moreira},
  journal  = {IEEE Transactions on Image Processing},
  title    = {Super-Resolution for Hyperspectral and Multispectral Image Fusion Accounting for Seasonal Spectral Variability},
  year     = {2020},
  pages    = {116-127},
  volume   = {29},
  doi      = {10.1109/TIP.2019.2928895},
}

@Article{Borsoi2021,
  author   = {Borsoi, Ricardo A. and Prévost, Clémence and Usevich, Konstantin and Brie, David and Bermudez, José C. M. and Richard, Cédric},
  journal  = {IEEE Journal of Selected Topics in Signal Processing},
  title    = {Coupled Tensor Decomposition for Hyperspectral and Multispectral Image Fusion With Inter-Image Variability},
  year     = {2021},
  number   = {3},
  pages    = {702-717},
  volume   = {15},
  doi      = {10.1109/JSTSP.2021.3054338},
}

@Article{Fu2022,
  author   = {Fu, Xiyou and Jia, Sen and Xu, Meng and Zhou, Jun and Li, Qingquan},
  journal  = {IEEE Transactions on Geoscience and Remote Sensing},
  title    = {Fusion of Hyperspectral and Multispectral Images Accounting for Localized Inter-Image Changes},
  year     = {2022},
  pages    = {1-18},
  volume   = {60},
  doi      = {10.1109/TGRS.2021.3124240},
}

@Article{Wang2023,
  author   = {Wang, Xiuheng and Borsoi, Ricardo Augusto and Richard, Cédric and Chen, Jie},
  journal  = {IEEE Transactions on Geoscience and Remote Sensing},
  title    = {Deep Hyperspectral and Multispectral Image Fusion With Inter-Image Variability},
  year     = {2023},
  pages    = {1-15},
  volume   = {61},
  doi      = {10.1109/TGRS.2023.3273118},
}

@Article{Simoes2014,
  author  = {Miguel Sim{\~o}es and Jos{\'e} M. Bioucas-Dias and Lu{\'i}s B. Almeida and Jocelyn Chanussot},
  journal = {IEEE Transactions on Geoscience and Remote Sensing},
  title   = {A Convex Formulation for Hyperspectral Image Superresolution via Subspace-Based Regularization},
  year    = {2014},
  pages   = {3373-3388},
  volume  = {53},
  url     = {https://api.semanticscholar.org/CorpusID:6406381},
}

@Article{Roger1996,
  author  = {R. E. Roger and John F. Arnold},
  journal = {International Journal of Remote Sensing},
  title   = {Reliably estimating the noise in AVIRIS hyperspectral images},
  year    = {1996},
  pages   = {1951-1962},
  volume  = {17},
  url     = {https://api.semanticscholar.org/CorpusID:128670666},
}

@Article{Shaw2003,
  author    = {Gary A. Shaw and Hsiao-hua K. Burke},
  journal   = {Lincoln Laboratory Journal},
  title     = {Spectral Imaging for Remote Sensing},
  year      = {2003},
  booktitle = {Lincoln Laboratory Journal},
  url       = {https://api.semanticscholar.org/CorpusID:360258},
}

@Article{Attouch2011,
  author    = {Attouch, Hedy and Bolte, Jérôme and Svaiter, Benar Fux},
  journal   = {Mathematical Programming},
  title     = {Convergence of descent methods for semi-algebraic and tame problems: proximal algorithms, forward–backward splitting, and regularized Gauss–Seidel methods},
  year      = {2011},
  issn      = {1436-4646},
  month     = aug,
  number    = {1–2},
  pages     = {91--129},
  volume    = {137},
  doi       = {10.1007/s10107-011-0484-9},
  publisher = {Springer Science and Business Media LLC},
}

@Article{Bolte2013,
  author    = {Bolte, Jérôme and Sabach, Shoham and Teboulle, Marc},
  journal   = {Mathematical Programming},
  title     = {Proximal alternating linearized minimization for nonconvex and nonsmooth problems},
  year      = {2013},
  issn      = {1436-4646},
  month     = jul,
  number    = {1–2},
  pages     = {459--494},
  volume    = {146},
  doi       = {10.1007/s10107-013-0701-9},
  publisher = {Springer Science and Business Media LLC},
}

@Article{Aiazzi2006,
  author   = {Aiazzi, B and Alparone, L and Baronti, S and Garzelli, A and Selva, M},
  journal  = {Photogrammetric engineering and remote sensing.},
  title    = {MTF-tailored Multiscale Fusion of High-resolution MS and Pan Imagery},
  year     = {2006},
  issn     = {0099-1112},
  month    = {May},
  number   = {5},
  pages    = {591—596},
  volume   = {72},
  url      = {http://europepmc.org/abstract/AGR/IND604805519},
}

@InProceedings{Kawakami2011,
  author    = {Kawakami, Rei and Matsushita, Yasuyuki and Wright, John and Ben-Ezra, Moshe and Tai, Yu-Wing and Ikeuchi, Katsushi},
  booktitle = {2011 IEEE Conference on Computer Vision and Pattern Recognition (CVPR)},
  title     = {High-resolution hyperspectral imaging via matrix factorization},
  year      = {2011},
  pages     = {2329-2336},
  doi       = {10.1109/CVPR.2011.5995457},
}

@Article{Yokoya2012,
  author   = {Yokoya, Naoto and Yairi, Takehisa and Iwasaki, Akira},
  journal  = {IEEE Transactions on Geoscience and Remote Sensing},
  title    = {Coupled Nonnegative Matrix Factorization Unmixing for Hyperspectral and Multispectral Data Fusion},
  year     = {2012},
  number   = {2},
  pages    = {528-537},
  volume   = {50},
  doi      = {10.1109/TGRS.2011.2161320},
}

@Article{Dong2016,
  author   = {Dong, Weisheng and Fu, Fazuo and Shi, Guangming and Cao, Xun and Wu, Jinjian and Li, Guangyu and Li, Xin},
  journal  = {IEEE Transactions on Image Processing},
  title    = {Hyperspectral Image Super-Resolution via Non-Negative Structured Sparse Representation},
  year     = {2016},
  number   = {5},
  pages    = {2337-2352},
  volume   = {25},
  doi      = {10.1109/TIP.2016.2542360},
}

@Article{Wei2015,
  author   = {Wei, Qi and Bioucas-Dias, José and Dobigeon, Nicolas and Tourneret, Jean-Yves},
  journal  = {IEEE Transactions on Geoscience and Remote Sensing},
  title    = {Hyperspectral and Multispectral Image Fusion Based on a Sparse Representation},
  year     = {2015},
  number   = {7},
  pages    = {3658-3668},
  volume   = {53},
  doi      = {10.1109/TGRS.2014.2381272},
}

@Article{Kolda2009,
  author   = {Kolda, Tamara G. and Bader, Brett W.},
  journal  = {SIAM Review},
  title    = {Tensor Decompositions and Applications},
  year     = {2009},
  number   = {3},
  pages    = {455-500},
  volume   = {51},
  doi      = {10.1137/07070111X},
  eprint   = {https://doi.org/10.1137/07070111X},
  url      = {https://doi.org/10.1137/07070111X},
}

@Article{Wang2004,
  author   = {Zhou Wang and Bovik, A.C. and Sheikh, H.R. and Simoncelli, E.P.},
  journal  = {IEEE Transactions on Image Processing},
  title    = {Image quality assessment: from error visibility to structural similarity},
  year     = {2004},
  number   = {4},
  pages    = {600-612},
  volume   = {13},
  doi      = {10.1109/TIP.2003.819861},
}

@Article{Wang2002,
  author   = {Zhou Wang and Bovik, A.C.},
  journal  = {IEEE Signal Processing Letters},
  title    = {A universal image quality index},
  year     = {2002},
  number   = {3},
  pages    = {81-84},
  volume   = {9},
  doi      = {10.1109/97.995823},
}

@Article{Dabov2007,
  author   = {Dabov, Kostadin and Foi, Alessandro and Katkovnik, Vladimir and Egiazarian, Karen},
  journal  = {IEEE Transactions on Image Processing},
  title    = {Image Denoising by Sparse 3-D Transform-Domain Collaborative Filtering},
  year     = {2007},
  number   = {8},
  pages    = {2080-2095},
  volume   = {16},
  doi      = {10.1109/TIP.2007.901238},
}

@Article{Zhang2017,
  author  = {Zhang, Kai and Zuo, Wangmeng and Zhang, Lei},
  journal = {IEEE Transactions on Image Processing},
  title   = {FFDNet: Toward a Fast and Flexible Solution for CNN based Image Denoising},
  year    = {2017},
  month   = {10},
  volume  = {PP},
  doi     = {10.1109/TIP.2018.2839891},
}

@InProceedings{Gu2014,
  author    = {Gu, Shuhang and Zhang, Lei and Zuo, Wangmeng and Feng, Xiangchu},
  booktitle = {2014 IEEE Conference on Computer Vision and Pattern Recognition},
  title     = {Weighted Nuclear Norm Minimization with Application to Image Denoising},
  year      = {2014},
  pages     = {2862-2869},
  doi       = {10.1109/CVPR.2014.366},
}

@Book{Bochnak1998,
  author    = {Bochnak, Jacek and Coste, Michel and Roy, Marie-Fran{\c{c}}oise},
  publisher = {Springer},
  title     = {Real Algebraic Geometry},
  year      = {1998},
  address   = {Berlin Heidelberg},
  doi       = {10.1007/978-3-662-03718-8},
}

@Article{Prevost2022,
  author   = {Pr\'{e}vost, Cl\'{e}mence and Borsoi, Ricardo A. and Usevich, Konstantin and Brie, David and Bermudez, Jos\'{e} C. M. and Richard, C\'{e}dric},
  journal  = {SIAM Journal on Imaging Sciences},
  title    = {Hyperspectral Super-resolution Accounting for Spectral Variability: Coupled Tensor LL1-Based Recovery and Blind Unmixing of the Unknown Super-resolution Image},
  year     = {2022},
  number   = {1},
  pages    = {110-138},
  volume   = {15},
  doi      = {10.1137/21M1409354},
  eprint   = {https://doi.org/10.1137/21M1409354},
  url      = {https://doi.org/10.1137/21M1409354},
}

@Article{Kanatsoulis2018a,
  author   = {Kanatsoulis, Charilaos I. and Fu, Xiao and Sidiropoulos, Nicholas D. and Ma, Wing-Kin},
  journal  = {IEEE Transactions on Signal Processing},
  title    = {Hyperspectral Super-Resolution: A Coupled Tensor Factorization Approach},
  year     = {2018},
  number   = {24},
  pages    = {6503-6517},
  volume   = {66},
  doi      = {10.1109/TSP.2018.2876362},
}

@Article{Dian2019,
  author   = {Dian, Renwei and Li, Shutao},
  journal  = {IEEE Transactions on Image Processing},
  title    = {Hyperspectral Image Super-Resolution via Subspace-Based Low Tensor Multi-Rank Regularization},
  year     = {2019},
  number   = {10},
  pages    = {5135-5146},
  volume   = {28},
  doi      = {10.1109/TIP.2019.2916734},
}

@Article{Dian2019a,
  author   = {Dian, Renwei and Li, Shutao and Fang, Leyuan},
  journal  = {IEEE Transactions on Neural Networks and Learning Systems},
  title    = {Learning a Low Tensor-Train Rank Representation for Hyperspectral Image Super-Resolution},
  year     = {2019},
  number   = {9},
  pages    = {2672-2683},
  volume   = {30},
  doi      = {10.1109/TNNLS.2018.2885616},
}

@Article{Kang2017,
  author   = {Kang, Xudong and Zhang, Xiangping and Li, Shutao and Li, Kenli and Li, Jun and Benediktsson, Jón Atli},
  journal  = {IEEE Transactions on Geoscience and Remote Sensing},
  title    = {Hyperspectral Anomaly Detection With Attribute and Edge-Preserving Filters},
  year     = {2017},
  number   = {10},
  pages    = {5600-5611},
  volume   = {55},
  doi      = {10.1109/TGRS.2017.2710145},
}

@Article{Akbari2010,
  author   = {Akbari, Hamed and Kosugi, Yukio and Kojima, Kazuyuki and Tanaka, Naofumi},
  journal  = {IEEE Transactions on Biomedical Engineering},
  title    = {Detection and Analysis of the Intestinal Ischemia Using Visible and Invisible Hyperspectral Imaging},
  year     = {2010},
  number   = {8},
  pages    = {2011-2017},
  volume   = {57},
  doi      = {10.1109/TBME.2010.2049110},
}

@Article{Akhtar2018,
  author   = {Akhtar, Naveed and Mian, Ajmal},
  journal  = {IEEE Transactions on Neural Networks and Learning Systems},
  title    = {Nonparametric Coupled Bayesian Dictionary and Classifier Learning for Hyperspectral Classification},
  year     = {2018},
  number   = {9},
  pages    = {4038-4050},
  volume   = {29},
  doi      = {10.1109/TNNLS.2017.2742528},
}

@Article{Peng2019,
  author   = {Peng, Jiangtao and Li, Luoqing and Tang, Yuan Yan},
  journal  = {IEEE Transactions on Neural Networks and Learning Systems},
  title    = {Maximum Likelihood Estimation-Based Joint Sparse Representation for the Classification of Hyperspectral Remote Sensing Images},
  year     = {2019},
  number   = {6},
  pages    = {1790-1802},
  volume   = {30},
  doi      = {10.1109/TNNLS.2018.2874432},
}

@Article{Zhang2017a,
  author     = {Zhang, Kai and Zuo, Wangmeng and Chen, Yunjin and Meng, Deyu and Zhang, Lei},
  journal    = {IEEE Transactions on Image Processing},
  title      = {Beyond a Gaussian Denoiser: Residual Learning of Deep CNN for Image Denoising},
  year       = {2017},
  issn       = {1057-7149},
  month      = jul,
  number     = {7},
  pages      = {3142–3155},
  volume     = {26},
  doi        = {10.1109/TIP.2017.2662206},
  issue_date = {July 2017},
  numpages   = {14},
  publisher  = {IEEE Press},
  url        = {https://doi.org/10.1109/TIP.2017.2662206},
}

@InProceedings{Yuan2020,
  author    = {Yuan, Yue and Wang, Qi and Li, Xuelong},
  booktitle = {2020 IEEE International Geoscience and Remote Sensing Symposium (IGARSS)},
  title     = {Hyperspectral and Multispectral Image Fusion Using Non-Convex Relaxation Low Rank and Total Variation Regularization},
  year      = {2020},
  pages     = {2683-2686},
  doi       = {10.1109/IGARSS39084.2020.9323227},
}

@Article{Rivenson2009,
  author  = {Rivenson, Yair and Stern, Adrian},
  journal = {Signal Processing Letters, IEEE},
  title   = {Compressed Imaging With a Separable Sensing Operator},
  year    = {2009},
  month   = {07},
  pages   = {449 - 452},
  volume  = {16},
  doi     = {10.1109/LSP.2009.2017817},
}

@Article{Debes2014,
  author   = {Debes, Christian and Merentitis, Andreas and Heremans, Roel and Hahn, Jürgen and Frangiadakis, Nikolaos and van Kasteren, Tim and Liao, Wenzhi and Bellens, Rik and Pižurica, Aleksandra and Gautama, Sidharta and Philips, Wilfried and Prasad, Saurabh and Du, Qian and Pacifici, Fabio},
  journal  = {IEEE Journal of Selected Topics in Applied Earth Observations and Remote Sensing},
  title    = {Hyperspectral and LiDAR Data Fusion: Outcome of the 2013 GRSS Data Fusion Contest},
  year     = {2014},
  number   = {6},
  pages    = {2405-2418},
  volume   = {7},
  doi      = {10.1109/JSTARS.2014.2305441},
}

@Article{CampsValls2006,
  author   = {Camps-Valls, G. and Gomez-Chova, L. and Munoz-Mari, J. and Vila-Frances, J. and Calpe-Maravilla, J.},
  journal  = {IEEE Geoscience and Remote Sensing Letters},
  title    = {Composite kernels for hyperspectral image classification},
  year     = {2006},
  number   = {1},
  pages    = {93-97},
  volume   = {3},
  doi      = {10.1109/LGRS.2005.857031},
}

@Article{Zhao2022,
  author   = {Zhao, Xueying and Bai, Minru and Sun, Defeng and Zheng, Libin},
  journal  = {SIAM Journal on Imaging Sciences},
  title    = {Robust Tensor Completion: Equivalent Surrogates, Error Bounds, and Algorithms},
  year     = {2022},
  number   = {2},
  pages    = {625-669},
  volume   = {15},
  doi      = {10.1137/21M1429539},
  eprint   = {https://doi.org/10.1137/21M1429539},
  url      = {https://doi.org/10.1137/21M1429539},
}

@Article{Chen2022,
  author   = {Chen, Yong and Zeng, Jinshan and He, Wei and Zhao, Xi-Le and Huang, Ting-Zhu},
  journal  = {IEEE Transactions on Geoscience and Remote Sensing},
  title    = {Hyperspectral and Multispectral Image Fusion Using Factor Smoothed Tensor Ring Decomposition},
  year     = {2022},
  pages    = {1-17},
  volume   = {60},
  doi      = {10.1109/TGRS.2021.3114197},
}

@Article{Xu2019,
  author   = {Xu, Yang and Wu, Zebin and Chanussot, Jocelyn and Wei, Zhihui},
  journal  = {IEEE Transactions on Image Processing},
  title    = {Nonlocal Patch Tensor Sparse Representation for Hyperspectral Image Super-Resolution},
  year     = {2019},
  number   = {6},
  pages    = {3034-3047},
  volume   = {28},
  doi      = {10.1109/TIP.2019.2893530},
}

@Article{Yang2025,
  author    = {Yang, Kunjing and Bai, Minru and Dian, Renwei and Lu, Ting},
  journal   = {Inverse Problems and Imaging},
  title     = {Subspace-based coupled tensor decomposition for hyperspectral blind fusion},
  year      = {2025},
  issn      = {1930-8345},
  number    = {3},
  pages     = {560--591},
  volume    = {19},
  doi       = {10.3934/ipi.2024045},
  publisher = {American Institute of Mathematical Sciences (AIMS)},
}

@Article{Landgrebe2002,
  author   = {Landgrebe, D.},
  journal  = {IEEE Signal Processing Magazine},
  title    = {Hyperspectral image data analysis},
  year     = {2002},
  number   = {1},
  pages    = {17-28},
  volume   = {19},
  doi      = {10.1109/79.974718},
}

@Article{Liu2024,
  author  = {Liu, Yuanye and Dian, Renwei and Li, Shutao},
  journal = {International Journal of Computer Vision},
  title   = {Low-Rank Transformer for High-Resolution Hyperspectral Computational Imaging},
  year    = {2024},
  month   = {08},
  pages   = {809-824},
  volume  = {133},
  doi     = {10.1007/s11263-024-02203-7},
}

@Article{Wang2023a,
  author   = {Wang, Xinying and Cheng, Cheng and Liu, Shenglan and Song, Ruoxi and Wang, Xianghai and Feng, Lin},
  journal  = {IEEE Transactions on Geoscience and Remote Sensing},
  title    = {SS-INR: Spatial-Spectral Implicit Neural Representation Network for Hyperspectral and Multispectral Image Fusion},
  year     = {2023},
  pages    = {1-14},
  volume   = {61},
  doi      = {10.1109/TGRS.2023.3317413},
}

@Article{Dian2018,
  author  = {Dian, Renwei and Li, Shutao and Guo, Anjing and Fang, Leyuan},
  journal = {IEEE Transactions on Neural Networks and Learning Systems},
  title   = {Deep Hyperspectral Image Sharpening},
  year    = {2018},
  month   = {02},
  pages   = {1-11},
  volume  = {PP},
  doi     = {10.1109/TNNLS.2018.2798162},
}

@Article{Luo2025,
  author   = {Sen Luo and Yurong Qian and Lu Bai and Yingying Fan and Yuanxu Wang and WeiQuan Kong},
  journal  = {Information Fusion},
  title    = {Deep learning-based hyperspectral and multispectral fusion techniques: Review, optimization, and perspectives},
  year     = {2025},
  issn     = {1566-2535},
  pages    = {103291},
  volume   = {124},
  doi      = {https://doi.org/10.1016/j.inffus.2025.103291},
  url      = {https://www.sciencedirect.com/science/article/pii/S1566253525003641},
}

@Article{Chen2021,
  author   = {Nan Chen and Lichun Sui and Biao Zhang and Hongjie He and Kyle Gao and Yandong Li and José {Marcato Junior} and Jonathan Li},
  journal  = {International Journal of Applied Earth Observation and Geoinformation},
  title    = {Fusion of Hyperspectral-Multispectral images joining Spatial-Spectral Dual-Dictionary and structured sparse Low-rank representation},
  year     = {2021},
  issn     = {1569-8432},
  pages    = {102570},
  volume   = {104},
  doi      = {https://doi.org/10.1016/j.jag.2021.102570},
  url      = {https://www.sciencedirect.com/science/article/pii/S0303243421002774},
}

@InProceedings{Dian2018a,
  author    = {Dian, Renwei and Li, Shutao and Fang, Leyuan and Bioucas-Dias, José},
  booktitle = {2018 IEEE International Geoscience and Remote Sensing Symposium (IGARSS)},
  title     = {Hyperspectral Image Super-Resolution via Local Low-Rank and Sparse Representations},
  year      = {2018},
  pages     = {4003-4006},
  doi       = {10.1109/IGARSS.2018.8519213},
}

@InProceedings{Eckardt2015,
  author    = {Eckardt, Andreas and Horack, John and Lehmann, Frank and Krutz, David and Drescher, Jürgen and Whorton, Mark and Soutullo, Mike},
  booktitle = {2015 IEEE International Geoscience and Remote Sensing Symposium (IGARSS)},
  title     = {DESIS (DLR Earth Sensing Imaging Spectrometer for the ISS-MUSES platform)},
  year      = {2015},
  pages     = {1457-1459},
  doi       = {10.1109/IGARSS.2015.7326053},
}

@InProceedings{Kaufmann2006,
  author    = {Kaufmann, H. and Segl, K. and Chabrillat, S. and Hofer, S. and Stuffler, T. and Mueller, A. and Richter, R. and Schreier, G. and Haydn, R. and Bach, H.},
  booktitle = {2006 IEEE International Symposium on Geoscience and Remote Sensing},
  title     = {EnMAP A Hyperspectral Sensor for Environmental Mapping and Analysis},
  year      = {2006},
  pages     = {1617-1619},
  doi       = {10.1109/IGARSS.2006.417},
}

@Article{Hilker2009,
  author   = {Thomas Hilker and Michael A. Wulder and Nicholas C. Coops and Julia Linke and Greg McDermid and Jeffrey G. Masek and Feng Gao and Joanne C. White},
  journal  = {Remote Sensing of Environment},
  title    = {A new data fusion model for high spatial- and temporal-resolution mapping of forest disturbance based on Landsat and MODIS},
  year     = {2009},
  issn     = {0034-4257},
  number   = {8},
  pages    = {1613-1627},
  volume   = {113},
  doi      = {https://doi.org/10.1016/j.rse.2009.03.007},
  url      = {https://www.sciencedirect.com/science/article/pii/S003442570900087X},
}

@Article{Emelyanova2013,
  author   = {Irina V. Emelyanova and Tim R. McVicar and Thomas G. {Van Niel} and Ling Tao Li and Albert I.J.M. {van Dijk}},
  journal  = {Remote Sensing of Environment},
  title    = {Assessing the accuracy of blending Landsat–MODIS surface reflectances in two landscapes with contrasting spatial and temporal dynamics: A framework for algorithm selection},
  year     = {2013},
  issn     = {0034-4257},
  pages    = {193-209},
  volume   = {133},
  doi      = {https://doi.org/10.1016/j.rse.2013.02.007},
  url      = {https://www.sciencedirect.com/science/article/pii/S0034425713000473},
}

@Article{Liu2019,
  author   = {Liu, Sicong and Marinelli, Daniele and Bruzzone, Lorenzo and Bovolo, Francesca},
  journal  = {IEEE Geoscience and Remote Sensing Magazine},
  title    = {A Review of Change Detection in Multitemporal Hyperspectral Images: Current Techniques, Applications, and Challenges},
  year     = {2019},
  number   = {2},
  pages    = {140-158},
  volume   = {7},
  doi      = {10.1109/MGRS.2019.2898520},
}

\end{document}